\documentclass[twoside]{article}
\usepackage{ecj,palatino,epsfig,latexsym}

\usepackage{graphicx} 
\usepackage[ruled,vlined]{algorithm2e}
\usepackage{algpseudocode}
\usepackage{colortbl}
\usepackage{arydshln}
\usepackage{euscript}
\usepackage{enumitem}
\usepackage{listings}
\usepackage{url}
\usepackage{booktabs}
\usepackage{siunitx}

\usepackage{amssymb, amsmath, amsthm, amsfonts}
\usepackage[square,sort,comma,numbers]{natbib}
\usepackage{diagbox}
\usepackage{pdflscape}
\usepackage{subcaption}

\begin{document}

\algnewcommand\algorithmicswitch{\textbf{switch}}
\algnewcommand\algorithmiccase{\textbf{case}}
\algnewcommand\algorithmicassert{\texttt{assert}}
\algnewcommand\Assert[1]{\State \algorithmicassert(#1)}%
\algdef{SE}[SWITCH]{Switch}{EndSwitch}[1]{\algorithmicswitch\ #1\ \algorithmicdo}{\algorithmicend\ \algorithmicswitch}%
\algdef{SE}[CASE]{Case}{EndCase}[1]{\algorithmiccase\ #1}{\algorithmicend\ \algorithmiccase}%
\algtext*{EndSwitch}%
\algtext*{EndCase}%

\ecjHeader{x}{x}{xxx-xxx}{201X}{IT Evolutionary Algorithm for Symbolic Regr.}{F. O. de Franca, G. S. I. Aldeia}

\title{\bf Interaction-Transformation Evolutionary Algorithm for Symbolic Regression}

\author{\name{\bf F. O. de Franca} \hfill \addr{folivetti@ufabc.edu.br}\\
          \addr{Center for Mathematics, Computation and Cognition, Heuristics, Analysis and Learning Laboratory, Federal University of ABC,
          Santo Andre, Brazil}
--\AND
         \name{\bf G. S. I. Aldeia} \hfill \addr{guilherme.aldeia@ufabc.edu.br}\\
          \addr{Center for Mathematics, Computation and Cognition, Heuristics, Analysis and Learning Laboratory, Federal University of ABC,
          Santo Andre, Brazil}
  }

\maketitle

\begin{abstract}
Interaction-Transformation (IT) is a new representation for Symbolic Regression that reduces the space of solutions to a set of expressions that follow a specific structure. The potential of this representation was illustrated in prior work with the algorithm called SymTree. This algorithm starts with a simple linear model and incrementally introduces new transformed features until a stop criteria is met. While the results obtained by this algorithm were competitive with the literature, it had the drawback of not scaling well with the problem dimension. This paper introduces an mutation only Evolutionary Algorithm, called ITEA, capable of evolving a population of IT expressions. One advantage of this algorithm is that it enables the user to specify the maximum number of terms in an expression. In order to verify the competitiveness of this approach, ITEA is compared to linear, nonlinear and Symbolic Regression models from the literature. The results indicate that ITEA is capable of finding equal or better approximations than other Symbolic Regression models while being competitive to state-of-the-art non-linear models. Additionally, since this representation follows a specific structure, it is possible to extract the importance of each original feature of a data set as an analytical function, enabling us to automate the explanation of any prediction. In conclusion, ITEA is competitive when comparing to regression models with the additional benefit of automating the extraction of additional information of the generated models.
\end{abstract}


%

\section{Introduction}

Regression analysis has the objective of describing the relationship between measurable variables~\citep{kass1990nonlinear}. A model generated by this task can be used to make predictions of unseen examples, to study a system's behavior or to calculate the statistical properties of such a system.

There is a wide range of algorithms proposed to create regression models with pre-specified forms ranging from the simpler \emph{linear model} to black-box models such as \emph{multi-layer perceptron}, with the latter example having the property of universal approximation~\citep{hornik1989multilayer}. These models usually have a fixed function form composed of the measurable variables and some \emph{adjustable parameters}, also called coefficients or weights. These parameters are optimized so that the model fits the original data being studied. 

Akin to this, \emph{Symbolic Regression} is the problem of searching for the optimal function form altogether with the adjusted parameters corresponding to the best fit. The general idea is to generate an explicit description of the system behavior that generalizes well with unseen data. This problem is often solved by an evolutionary approach called \emph{Genetic Programming}.

When applying Genetic Programming to Symbolic Regression, the function form is commonly (but not necessarily) represented by an \emph{expression tree}, with each node containing a constant, measured variable, or a function of any arity. This representation has the advantage of comprising the whole set of possible function forms. As a side effect, this set also includes redundant expressions and, with that, the possibility of creating  unnecessarily large functions. Also, in some situations the parameters adjustment step can require an expensive nonlinear optimization procedure.

Recently, an alternative representation called \textbf{Interaction-Transformation} was introduced in~\cite{de2018greedy} within the context of Symbolic Regression. The basic idea is that the search space contains only mathematical expressions described as an affine combination of nonlinear transformations of different interactions between the original variables. As a result, this representation creates a new set of transformed variables that hopefully expresses a linear relationship with the target variable.

Also in~\cite{de2018greedy} a new greedy algorithm called \textbf{SymTree} was introduced and tested on a set of low-dimension benchmark functions. In this benchmark, SymTree was capable of finding better approximations than some Genetic Programming approaches and traditional regression algorithms. The author noted that the main downside of this algorithm is that it does not scale well with the problem dimension, but that this could be alleviated by using search meta-heuristics such as \emph{Evolutionary Algorithms}.

In this paper we introduce an Evolutionary Algorithm for Symbolic Regression that evolves an Interaction-Transformation expression, called \textbf{Interaction-Transformation Evolutionary Algorithm} (ITEA). In order to validate this approach, the algorithm will be applied to a set of real-world benchmarks commonly used in Genetic Programming literature~\citep{martins2018solving,fracasso2018multi} and compared to the performance of traditional and state-of-the art algorithms.

This paper is organized as follows.Section~\ref{sec:background} gives a brief background about the regression problem. After that, Section~\ref{sec:it} details the Interaction-Transformation representation together with a brief explanation of SymTree and what has been done so far. In Section~\ref{sec:itea}, we introduce the Interaction-Transformation Evolutionary Algorithm giving sufficient implementation details. Following, in Section~\ref{sec:methods}, we explain the methodology adopted in this paper and, in Section~\ref{sec:results}, we report and discuss the experimental results obtained with ITEA. Finally, in Sectioonn~\ref{sec:conclusion}, we conclude this paper with some final comments regarding the results and perspectives for future research.

\section{Background}
\label{sec:background}

The regression problem can be formalized~\footnote{In this paper we will adopt uppercase letters for matrices, lowercase bold or Greek letters for vectors and lowercase letters for scalars.} as follows, given a set of $n$ $d$-dimensional samples in the form of a matrix $X \in \mathbb{R}^{n \times d}$, with each $\mathbf{x_i} \in \mathbb{R}^d$  denoted as an independent variable, and a set of scalars in the form of a vector $\mathbf{y} \in \mathbb{R}^{n}$, named dependent variables, our goal is to find a mapping $f: \mathbb{R}^{d} \rightarrow \mathbb{R}$ that maps any sampled $\mathbf{x} \in X$ to the corresponding measured $y \in \mathbf{y}$. 
In other words, we wish to obtain $f(\mathbf{x_i}) \approx y_i$ for each $i = 1 \ldots n$. 

This is often performed by parametric models in which the function form is defined \emph{a priori} and the mapping is found by adjusting a set of free parameters $\beta$. In such case, the function becomes $f(\mathbf{x}, \beta)$ with the goal of finding the optimal beta ($\beta^{*}$) that minimizes the approximation error of the model.
A simple example of a parametric model is the \textbf{linear regression} that assumes a function of the form:

\begin{equation}
f(\mathbf{x}, \mathbf{\beta}) = \mathbf{\beta} \cdot \mathbf{x} + \beta_0,
\label{eq:linreg}
\end{equation}

\noindent this model assumes that the relationship between independent and dependent variables is linear, an assumption that does not hold with many data sets. Despite that, this model is frequently used because of its descriptiveness that it makes it easier to understand how the variables interact with the system being studied.

Another well-known example of a parametric model is the one described as a \textbf{Multi-layer Perceptron} (MLP), that may assume different function forms depending on the topology like, for example:

\begin{equation*}
f(\mathbf{x}, \beta, \gamma) = \sum_i{\gamma_i \cdot g(\beta_i \cdot \mathbf{x})},
\end{equation*}

\noindent where $g(.)$ is a function called \textbf{activation function}. The activation function is usually a non-linear sigmoid function.
This model has the property of being an universal approximator~\citep{hornik1989multilayer}. Given the correct dimensions and values for $\beta$ and $\gamma$, it is possible to approximate any given function with a small error $\epsilon$.

The models that are either difficult or impossible to interpret are called \textbf{black box} because they do not make the relationships between variables explicit. One example belonging to this class of models is the \textbf{Multi-layer Perceptron} explained above. Even though there are some methods capable of extracting an estimated interpretation of any model~\cite{ribeiro2016model}, they are just local estimations of the generated model.


Different from these approaches, \textbf{Symbolic Regression}~\citep{koza1994genetic, langdon1999size, poli2008field, davidson2003symbolic} searches for the \emph{function form} that best fits the input data, introducing flexibility into the model. A common algorithm employed to solve this problem is the \textbf{Genetic Programming}~\citep{koza1994genetic, poli2008field,icke2013improving}, an Evolutionary Algorithm proposed to evolve computer programs. 
A common representation when using Genetic Programming with Symbolic Regression is the Expression Tree. The main advantage of this representation is that it can represent any function from the search space but, on the other hand, they are prone to \emph{bloat}. If not properly controlled, this can lead to very large expressions or chaining of nonlinear functions, making the expression harder to analyse and possibly limiting its generalization capabilities. 

One solution to this problem is to incorporate a \emph{parsimoony} measure to the fitness function, either in the form of penalization or in a multi-objective approach~\citep{vladislavleva2009order, fracasso2018multi}. Another solution is to represent each individual as a set of smaller trees combined by a regularized linear regression, with the objective of using only a subset of the trees to compose the final model. This approach is called Multiple Regression GP (MRGP)~\citep{arnaldo2014multiple} and its current state-of-the-art is the recently proposed Feature Engineering Automation Tool (FEAT)~\citep{cava2018learning, LaCava:2019:SVO:3321707.3321776}.

Another possible solution is to create a constrained representation powerful enough to fit nonlinear relations but restrictive to the point of making it impossible to have some complex constructs. A recent example of this approach, which is extended in this paper, is the \textbf{Interaction-Transformation} representation~\citep{de2018greedy, aldeia2018lightweight}. This representation together with the search heuristic introduced in that paper will be explained in the next section.

\section{Interaction-Transformation Symbolic Regression}
\label{sec:it}

Interaction-Transformation (IT) is a representation proposed in \cite{de2018greedy} that disallows \emph{complex} expressions and avoids some redundancy in the search space.

Given the definitions of a transformation function $t: \mathbb{R} \rightarrow \mathbb{R}$, as any univariate function, and an interaction function $p: \mathbb{R}^{d} \rightarrow \mathbb{R}$ for a $d$-dimensional space and described as:

\begin{equation}
p(\mathbf{x}) = \prod_{i=1}^{d}{x_i^{k_i}},
\end{equation}

\noindent with $k_i \in \mathbb{Z}$ called the \textbf{strength} of the interaction. We can define an IT expression for regression as a function with the form:

\begin{equation}
f(\mathbf{x}) = w_0 + \sum_{i}{w_i \cdot (t_i \circ p_i) (\mathbf{x})},
\label{eq:appfunction}
\end{equation}

\noindent where $w_i \in \mathbb{R}$ is the $i$-th coefficient of a linear combination, hereafter referred to as \textbf{weight}

This representation has the advantage of restricting the search space to expressions such as 

\begin{equation}
f(\mathbf{x}) = 3.5\sin{(x_1^2 \cdot x_2)} + 5\log{(x_2^3/x_1)},    
\label{eq:example}
\end{equation}

\noindent while not allowing more complicated function forms such as those with function chaining, like $f(\mathbf{x}) = \tanh{(\tanh{(\tanh{(\mathbf{w} \cdot \mathbf{x})})})}$. Notice that this sums up as a compromise between the simplicity of an expression generated by this representation and the completeness provided by the tree representation. In the event that we have data generated by an expression outside our representation capabilities, we can still approximate it with a polynomial interpolation if the training data is representative~\cite{stone1948generalized}.

An IT expression can be represented computationally as a triple $(T, F, W)$  encapsulating a matrix of strengths for each term and every variable, the list of functions to be applied to each term and a list of weights. For example, the expression in Eq.~\ref{eq:example} can be represented as $([[2,1], [-1,3]], [\sin, \log], [3.5, 5.0])$.

\subsection{Symbolic Regression Search Tree}

In the same paper~\citep{de2018greedy}, the Symbolic Regression Search Tree algorithm (SymTree) was introduced as an algorithm to benchmark the potential of the IT expressions. The procedure is similar to a breadth-first search which starts with a queue initialized with a state representing a linear model (such as in Eq.~\ref{eq:linreg}). At every step, the algorithm takes the first expression from the queue and expands it enumerating all neighboring states. Finally, it performs a filtering procedure to reduce the number of new states and add them to the queue. This algorithm stops until there are no more expressions in the queue or a limited number of iterations is reached. 

In order to generate the neighbor states of a given expression the algorithm first creates a set of candidates terms either by combining two terms from the parent expression or as a copy of one of its terms replacing the transformation function with a different one. 

The combination of terms follows a procedure called \textbf{positive} or \textbf{negative} interaction. In regression analysis, the variables set is often expanded by inserting interaction of the variables. For example, given two variables $x_1$ and $x_2$ they can create interactions such as $x_1 \cdot x_2$, $x_1^2 \cdot x_2^3$ or $x_1 \cdot x_2^{-1}$, just to show a few.

The \textbf{positive interaction} of two interactions is the multiplication of these interactions, so if we have:

\begin{equation*}
    p_1(x) = \prod_{i=1}^{d}{x_i^{k^1_i}}, \quad  p_2(x) = \prod_{i=1}^{d}{x_i^{k^2_i}},
\end{equation*}

\noindent the positive interaction would be

\begin{equation*}
    p_{1,2}(x) = \prod_{i=1}^{d}{x_i^{k^1_i+k^2_i}},
\end{equation*}

\noindent analogously, the \textbf{negative interaction} of two interactions is the division of the first interaction by the second one.
This is equivalent to a pairwise addition or subtraction of two lines of the terms matrix $T$. For example, a combination of the terms $x_1^2 \cdot x_2$ ($[2, 1]$) and $x_1^{-1} \cdot x_2^{3}$ ($[-1, 3]$) could generate the interaction $x_1 \cdot x_2^{4}$ ($[1, 4]$). The new transformation function is just the exchange of one element of the list $F$ to another one, for example, if $F = [\sin, \log]$, it could generate $F = [\cos, \log]$.

After generating the candidate set of new terms $\{t_1, t_2, t_3\}$, the power set of those terms is generated as $\mathbb{P} = \{\emptyset, \{t_1\}, \{t_2\}, \{t_3\}, \{t_1, t_2\}, \{t_1, t_3\}, \{t_2, t_3\}, \{t_1, t_2, t_3\}\}$. This power set is split into two subsets: the set of solutions that improves the parent solution, called \emph{candidate solutions}, and the complement of this set, called \emph{terminal states} that no longer will be expanded. Every set $p_i \in \mathbb{P}$ that is a subset of any set $p_j \in \mathbb{P}$ is removed from the set. Finally, each remaining set will become a new node of this search tree.

This search algorithm presented better results when compared to some variations of Genetic Programming algorithms and competitive results against black-box models while maintaining the length of the expression at a minimum. One noted drawback of this algorithm is that, in the worst case, the number of child nodes will grow exponentially with respect to the problem dimension, making the algorithm inappropriate for high-dimensional data sets. For more details regarding this algorithm, we refer the reader to the original paper~\citep{de2018greedy}.

\section{Interaction-Transformation Evolutionary Algorithm}
\label{sec:itea}

The Interaction-Transformation Evolutionary Algorithm (ITEA) proposed in this paper follows a  mutation-based evolutionary algorithm that starts with a randomly generated population of solutions, creates new solutions by applying a mutation operator as described in Section~\ref{sec:mutation}, and samples the population for the next generation with a selection procedure proportional to their fitness.
Each of these steps will be described in the following subsections together with some technical details of the implementation for the sake of reproducibility.


Many evolutionary algorithms also relies on a recombination operator called \emph{crossover}. But, in order to be effective, this operator should have a clear semantic meaning for the solution representation~\cite{eiben2002evolutionary}. In Symbolic Regression, classical crossover operators do not have such semantic meaning and thus require more elaborated algorithms~\cite{poli1998schema, ruberto2020sgp}. But, as we will show in Section~\ref{sec:results}, the different mutation operators proposed in this paper are enough to keep the algorithm competitive with the state-of-the-art. 

\subsection{Representation}

Each solution is the encapsulation of an IT expression together with the parameters obtained by the linear regression of the terms and the information of fitness value for that particular individual. As stated before, the expression can be numerically represented since each term can be described by the strength of its interactions and an id of the transformation function that should be applied.

As such, each individual contains a list of terms, with each term represented by a list of interactions strength~\footnote{This can be implemented more efficiently as an associative array ignoring the zero-valued strengths.}, a list of transformation functions, a list of weights, a variable representing the intercept of the linear regression, and another variable representing the fitness value, so that the data structure would look like:

\begin{lstlisting}{C}
Individual = {
  terms     :: [[Int]],
  funs      :: [Function Ids],
  weights   :: [Double],
  intercept :: Double,
  fitness   :: Double
}
\end{lstlisting}

For example, one individual representing Eq.~\ref{eq:example} would be encapsulated as:

\begin{lstlisting}{C}
Individual {
  terms     = [[2, 1], [-1, 3]];
  funs      = [sin, log];
  weights   = [3.5, 5.0];
  intercept = 0.0;
  fitness   = 5.29e-3;
}
\end{lstlisting}

\noindent assuming the fitness function to be the root mean squared error for the data $X = [[1,1], [0.1, 0.2]], y = [2.94, -3.14]$.

The procedure to create a random individual starts by first creating an $n \times d$ random integer matrix of strengths coefficients, with $n$ sampled from an uniform distribution, thus generating the \emph{terms} matrix. Then, it chooses $n$ random transformation functions, with replacement, to form the \emph{funs} list. In the last step, the values of \emph{weights, intercept, fitness} are calculated by first fitting a linear model of choice and then calculating the fitness with the chosen error (minimization) or performance (maximization) measure.

\subsection{Initial Population}

In order to create a random initial population, each individual is created following this set of rules:

\begin{itemize}
    \item The IT expression must have at least one term and at most $k$ terms.
    \item The strength of each interaction for every term must be within a range $[lb, ub]$.
    \item The list of transformation functions must be sampled from a provided list of available functions.
\end{itemize}

These rules have the intention of controlling the complexity of the generated expressions as well as integrate some prior knowledge or expectation the practitioner may have.

The first rule prevents the creation of the trivial expression with only the intercept and, also, of very large expressions. This parameter controls how much the practitioner is willing to sacrifice generalization power for simplicity.

With the second rule we can control the complexity of the expression by avoiding higher order polynomial terms that may limit the interpretability of the final expression due to high correlation between them.

Common to any Symbolic Regression algorithm, the last rule is just a choice given to the practitioner to choose a set of functions that makes sense in their context. It is worth noticing that a safe version of the function set is not required since the algorithm will simply discard any infeasible term automatically without compromising the whole expression. 

\subsection{Mutation}
\label{sec:mutation}

During the mutation step, each solution is modified by applying one out of six possible mutation algorithms chosen completely at random with the same probability. Each mutation changes one aspect of the IT expression:

\begin{itemize}
    \item \textbf{Drop term mutation:} removes one random term from the expression. This is only applied if there is a minimum number of terms in the current expression in order to avoid trivial expressions.
    \item \textbf{Add term mutation:} adds a new random term to the expression. This is only applied if there is at most a certain number of terms in the current expression in order to avoid \emph{bloat}.
    \item \textbf{Replace interaction mutation:} replaces a random interaction strength of a random term from the expression.
    \item \textbf{Positive interaction mutation:} replaces a random term from the expression with the positive interaction with another random term. This changes the interaction strength by performing an element-wise addition of both strength lists.
    \item \textbf{Negative interaction mutation:} replaces a random term from the expression with the negative interaction with another random term. This changes the interaction strength by performing an element-wise subtraction of both strength lists.
    \item \textbf{Replace transformation mutation:} replaces the transformation function of a random term from the expression.
\end{itemize}

The \emph{add} and \emph{drop} term mutations are responsible for regulating the length of the expression as well as adding novelty to the expression. The \emph{positive} and \emph{negative} mutations work as a local search by exploring the neighborhood of terms, similarly to the SymTree algorithm. The replacement of the interaction strengths also creates another way to explore the neighborhood of a solution  while preserving the transformation function. The replace transformation allows to test new transformation functions for a given interaction.

Overall, the idea is that the application of the mutation operator creates new terms to the expression that are afterwards regulated by the coefficient adjustment. This can lead to three different behaviors: i) the new term is ignored by setting the coefficient to zero; ii) the new term improves the fitness of the expression; iii) one or more terms of the expression are replaced by this new term when their coefficients are adjusted to zero. The only exception is the \emph{drop} mutation in which the sole purpose is to remove terms that either do not contribute to the expression or increase the approximation error. 

After applying the mutation, the new expression is fitted again using the linear regression algorithm chosen by the user, replacing the original list of weights, intercept and fitness. 
Notice that when a new term is added, it may have no effect on the fitness if the corresponding weight is equal to zero. Such terms can be eliminated from the expression as a simplification step but, in this paper, we have chosen not to in order to keep them as a building block for new terms.

\subsection{Selection}

The selection scheme should sample the solutions from the current and the mutated population forming a new population of solutions.
This sampling should favor the fittest individuals but should also give a chance for diversifying the population. The selection step can be one of the commonly used in evolutionary algorithms such as \emph{Elitism}, \emph{Total Replacement}, \emph{Roulette Wheel Selection} or \emph{Tournament Selection}.

\subsection{Evaluation of exceptions}

Genetic Programming practitioners traditionally choose a function set and operators that does not allow discontinuity. For example, instead of a division operator, the algorithm can use a safe division operator that returns constant value when a division-by-zero error occurs. This has the advantage of not completely discarding problematic functions and operators but, on the other hand, it can worsen the convergence of the algorithm since a large piece of the expression tree may actually be an evaluation error replaced by that constant value. In some cases, there are also some alternative operators that are total functions and approximates to the original operator, such as the analytical quotient~\cite{ni2012use} or the composition of square root with absolute value.

In ITEA, since each term of the IT expression is evaluated independently of one another before fitting the linear coefficients, we can simply discard any terms that generates any error within the training data set. If the training set is representative, we can use this approach to generate valid models 
without any concerns about discontinuity. But, assuming the evaluation error only happens on unseen samples, we can apply the protected operator \emph{a posteriori}. In this scenario, the result would be equivalent to applying the protected operator in the training set, since the protection mechanism would not be used during the search.

\subsection{User-defined parameters}

ITEA has a set of user-defined parameters, most of them common to Genetic Programming algorithms, to adjust some aspects of the algorithm behavior. These parameters should be set in order to better reflect the expectations for the final expression and, also, to reach a compromise between quality of solution and computational performance:

\begin{itemize}
    \item \textbf{Population size ($pop$):} the size of the population. The higher this value, the larger the exploration of the search space but with a compromise of computational performance.
    \item \textbf{Set of Transformation functions ($funcs$):} the set of functions to be considered when creating an expression. This is domain-specific and should  reflect the properties of the studied data set. If there is no prior knowledge of the data, the algorithm should still work with a large and common set of functions, even though the convergence rate may become slower.
    \item \textbf{Stop criteria ($stop$):} when the algorithm should stop iterating. The criteria should allow the algorithm to stop whenever the population converges to a single solution or for as much computational budget they have.
    \item \textbf{Maximum number of terms ($n\_terms$):} maximum number of terms when creating a random solution. A smaller number favors simpler solutions, but limits the search space.
    \item \textbf{Range of strength ($lb, ub$):} the range of the interaction strength when creating a random solution. Similar to the previous parameter, this controls the simplicity of the initial solutions and limits the search space.
    \item \textbf{Minimum length for \emph{drop} mutation ($min\_drop$):} the minimum number of terms necessary to apply the \emph{drop} mutation. This parameter avoids the creation of trivial solutions with a small number of terms. Notice that it is still possible to create a solution with less than $min\_drop$ since the weight of the terms can be set to zero during the fitting step.
    \item \textbf{Maximum length for \emph{add} mutation ($max\_add$):} the maximum number of terms to allow the application of the \emph{add} mutation. The opposite of the previous parameter, it avoids the creation of large expressions.
    \item \textbf{Linear Regression fitting algorithm ($model$):} the algorithm that should be used to fit the linear coefficients of the expression.
    \item \textbf{Fitness measure ($fitness$):} the objective-function used to evaluate the expression.
\end{itemize}

\section{Experimental Methods}
\label{sec:methods}

\begin{table}[t!]
\caption{Real-world data sets used in the experiments.}
    \label{tab:datasets}
    \centering
    \begin{tabular}{ccc}
        \toprule
        \textbf{Data set} & \textbf{Samples} & \textbf{Features} \\
        \midrule
        Airfoil  & $1,503$ & $5$ \\
        Concrete & $1,030$ & $8$ \\
        Energy Cooling & $768$ & $9$\\
        Energy Heating & $768$ & $9$\\
        Geographical Original of Music & $1,059$ & $117$ \\
        Tecator & $240$ & $124$\\
        Tower Data & $4,999$ & $25$\\
        Wine Red & $1,599$ & $10$\\
        Wine White & $4,898$ & $10$\\
        Yacht & $308$ & $6$\\
        \bottomrule
    \end{tabular}
    
\end{table}

In this section we assess the performance of the proposed algorithm when compared to a representative set of Symbolic Regression algorithms and a set of other regression models. The goals of this experimental setup are to verify if ITEA improves upon its predecessor, SymTree; how it compares to state-of-the-art Symbolic Regression algorithms; and where it stands when compared to commonly used regression models performance wise. Additionally, we will verify whether each mutation operator contributes to the performance of the algorithm or not.

The performance of the algorithms will be measured on $8$ different real-world data sets taken from~\cite{lichman2013uci, albinati2015effect,Olson2017PMLB}. Each data set was split into five folds for a cross-validation procedure. In this work, we use the Root Mean Squared Error as the evaluation metric, calculated by:

\begin{equation}
    RMSE(\mathbf{y}, \mathbf{\hat{y}}) = \sqrt{\frac{1}{n}\sum_{i=1}^{n}{(y_i - \hat{y}_i)^2}},
    \label{eq:rmse}
\end{equation}

\noindent where $\mathbf{\hat{y}}$ is the vector of predicted values for each sample.
The name, number of samples and number of features of each data set are given in Table~\ref{tab:datasets}. As we can see, the chosen data sets have a varying number of samples and features. 

\subsection{Contribution of each Mutation Operator}

As a first experiment, we will assess the contribution of each mutation operator of ITEA to the search process. So, we have run several versions of ITEA with a different subset of the mutation operators. Initially, we tested the subsets \{Add Term, Drop Term\}, \{Replace Interaction\}, \{Positive Interaction, Negative, Interaction\}, \{Replace Transformation\} individually. After that, we calculated the average rank of each group and incrementally merged the best set with the other sets until no more improvement is observed. 

For this experiment we performed $100$ ($20$ times for each fold) independent runs for each dataset with a fixed configuration of population size and number of iterations both set to $100$, maximum number of $15$ terms, the range of the strength coefficients set to $(-3, 3)$ and the transformation function set $\{ id, sin, cos, tanh, \sqrt{|.|}, log, exp\}$. Notice that the results of this experiment cannot be used to compare with other approaches from the literature and, as such, we will only report the average rank of each tested version, to avoid any biased conclusion.

\subsection{Comparison with other approaches}

For our second experiment, we will compare the results obtained by ITEA with different regression algorithms from the literature. In this experiment, each algorithm adjusts the model to the training set and then, the best model is applied to the test set. This procedure is repeated $6$ times per fold for the stochastic algorithms, totaling $30$ runs for each data set. Also, it should be noted that these data sets were not processed in any way during the experiments, so that every algorithm received the very same data as their input. All of the experiments were statistically tested with a pairwise Wilcoxon Rank Sum test with Bonferroni adjustment. The p-values are reported in Table~\ref{tab:pvals}.

The obtained results are compared with some standard linear regression algorithm, some non-linear machine learning models, and with a set of Symbolic Regression algorithms as well as with the original IT-based algorithm. In short, the choice of algorithms together with their corresponding abbreviated names (in parentheses) are:

\begin{itemize}
    \item \textbf{Ordinary Least Square with Ridge regularization (Ridge):} Linear model solved with an $l_2$ regularization as the minimization of $\|X \cdot \beta - \mathbf{y} \|^2_2 + \lambda \|\beta\|^2_2$. The coefficients $\beta$ are given by the closed form solution $\beta = (X^TX + \lambda I)^{-1}X^T\mathbf{y}$.
    \item \textbf{Coordinate Descent with Lasso regularization (Lasso):} Linear model solved with an $l_1$ regularization as the minimization of $\|X \cdot \beta - \mathbf{y} \|^2_2 + \lambda \|\beta\|_1$. The coefficients $\beta$ are given by the Coordinate Gradient Descent method.
    \item \textbf{Least Angle Regression (LARS):} the same model as Lasso but with the coefficients adjusted by a forward stagewise selection linear regression that estimates the optimal coefficients for different values of $\lambda$.
    \item \textbf{Elastic Net (ElNet):} Linear model combining $l_1$ and $l_2$ regularization with the coefficients adjusted by the Coordinate Gradient Descent method.
    
    \item \textbf{k-Nearest Neighbors (kNN):} a non-parametric model~\cite{dasarathy1991nearest} that estimates the value of a new point as the average of its $k$ closest neighbors. This algorithm allows only for a local interpretation.
    
    \item \textbf{Decision Tree Regressor (Tree):} this model is represented as a decision tree with each node containing a predicate in the form $x_i \leq \tau$, with $x_i$ being a predictor and $\tau$ a threshold constant. Every predicate splits the training data into disjoint subsets in such a way that each split minimizes a loss function. Whenever the size of the subsets generated by the splits is smaller than a percentage of the original set, the splitting stops and those generated nodes become leaf nodes. The target value for a new sample is estimated by traversing the tree through the edges where the predicate is true for that particular sample then, when it reaches a leaf node, the target value is calculated as the average value for that particular subset of the training data~\cite{utgoff1989incremental}.
    
    \item \textbf{Random Forest (Forest):} ensemble of $n$ Decision Trees using bagging. Notice that this particular model is considered a black-box model unlike all of the other methods tested on this paper~\cite{liaw2002classification}.
    
    \item \textbf{Symbolic Regression Tree (SymTree):} the algorithm introduced with the Interaction-Transformation representation and explained in Sec.~\ref{sec:it}. The only parameters of  this algorithm are the significance threshold that removes predictors with an absolute value smaller than this threshold and the maximum number of iterations. Notice that this last parameter should be set to a small value in order to avoid an exponential growth of the size of the expressions.
    \item \textbf{Canonical Genetic Programming (GP):} a canonical implementation of Genetic Programming with standard crossover and mutation operators that replaces a given node with a random subtree, replaces a single node with a new symbol, or reduces the height of a subtree by exchanging it with a random child subtree~\citep{poli2008field}. The initialization of the population is performed by the ramped half-and-half tree generator~\citep{langdon1999size}.

    \item \textbf{Differentiable Cartesian Genetic Programming (dCGP):} in Cartesian Genetic Programming~\citep{miller2008cartesian} the expressions are encoded as a list of integers that represents a flattened graph structure. The \emph{unflattened} structure is a graph with a grid structure composed of a number of rows and columns, as per suggestion of~\cite{miller2019cartesian}, the number of rows is usually set as $1$. The fixed-length segments of this list represents a connection of a set of predictors to a function. Differentiable Cartesian Genetic Programming~\cite{izzo2017differentiable} applies automatic differentiation to adjust the constant coefficients of the decoded expression. The expressions are evolved by a mutation-only evolutionary algorithm.
    \item \textbf{Geometric Semantic Genetic Programming (GSGP):} in Geometric Semantic Genetic Programming~\cite{moraglio2012geometric} the evolution of symbolic expressions are given by the Geometric Semantic crossover that combine two parent trees $T_1, T_2$ as a weighted average $(T_r \cdot T_1) + ((1 - T_r) \cdot T_2)$, where $T_r$ is a random function with codomain in the range of $[0,1]$. Similarly, the Geometric Semantic Mutation changes a tree $T$ with the expression $T + ms \cdot (T_{r1} -T_{r2})$, with $ms$ being a random mutation step in the range $[0,1]$, and $T_{r1}, T_{r2}$ are random functions with the same characteristics as $T_r$.
    \item \textbf{Feature Engineering Automation Tool (FEAT):} in~\citep{cava2018learning,LaCava:2019:SVO:3321707.3321776} in Feature Engineering Automation Tool (FEAT) an expression is given by an affine combination of smaller expression trees (sub-trees). The initial population is composed of one individual representing a trivial linear model and the remainders as an affine combination of random expression trees of random depths (up to a maximum depth). Afterwards, a standard evolutionary algorithm procedure is applied with one of the following mutation operators: \emph{point mutation}, where a single node is replaced by another with the same arity;  \emph{insert mutation}, that replaces a node with a depth $1$ random sub-tree; \emph{delete mutation}, that removes a sub-tree or replaces one by a singleton tree containing one of the original predictors; \emph{insert/delete dimension}, that inserts or deletes a sub-tree. Additionally, the parents can undergo one of two crossover operations: \emph{sub-tree crossover}, that exchange a sub-tree from one parent with the sub-tree of another; \emph{dimension crossover}, swaps two features from different parents. Besides the functions and operators commonly used in the literature, the authors also included functions often used as activation functions of Neural Network and boolean functions. Despite the similarities with the IT representation, in FEAT each term has a free form limited by a maximum depth and number of original predictors involved, conversely in IT, each term has a pre-defined structure.
\end{itemize}

\begin{table}[t!]
\centering
\caption{Set of parameters values used during the Grid Search.}
\begin{tabular}{ll}
\toprule
Algorithm & Parameters \\
\midrule
ITEA & population $= \{100, 250, 500\}$, \\
     & generations $= 100,000 /$ population \\
     & $min\_drop = 2$ \\
     & $max\_add = \{10, 15\}$ \\
     & strength range $= \{(-3,3), (-2,2)\}$ \\
\midrule
SymTree & significance threshold $= \{1e-3, 1e-4, 1e-5, 1e-6\}$\\
        & iterations $= \{1, 2, 3, 4, 5\}$ \\
        & (or a maximum of $1$ hour of execution time)\\
\midrule
GP, GSGP, FEAT   & population $= \{100, 250, 500\}$, \\
           & generations $= 100,000 /$ population \\
           & crossover rate $= \{0.2, 0.5, 0.8\}$ \\
           & mutation rate $= 1 - $ cross. rate \\
           & feedback (feat) $= \{0.2, 0.5, 0.8\}$ \\
\midrule
dCGP & rows $= 1$ \\
     & columns $= \{100, 200, 500\}$\\
     & generations $= 100,000 /$ columns \\
\midrule
$k$NN & $k = \{1, 2, 3, \ldots, 0.3 \cdot |Tr|\}$,\\
      & where $|Tr|$ is the size of the training set \\
\midrule
Tree, Forest & loss function $=$ mean squared error, \\
     & minimum split $= \{1\%, 5\%, 10\%\}$ \\
     & number of estimators (forest) $n = \{100, 200, 300\}$  \\
\midrule       
ElNet, Lars, Lasso, Ridge & $\lambda = \{0.001, 0.01, 0.1, 1, 10\}$ \\
\bottomrule
\end{tabular}
\label{tab:params}
\end{table}

For the non-GP based algorithms we use the implementations provided by the Scikit-Learn Python library~\citep{pedregosa2011scikit} version $0.20$. We use \emph{GPLearn} Python library~\footnote{https://gplearn.readthedocs.io/en/stable/} for the canonical GP algorithm and \emph{dCGP}~\citep{dario_izzo_2019_3544905} library provided by the authors of the algorithm. For GSGP~\footnote{http://gsgp.sourceforge.net/?page\_id=35}, we have used the implementation published in~\cite{castelli2019gsgp} with adaptations of our own~\footnote{https://github.com/gAldeia/SOFTX\_2019\_170} to work with the same function as the other implementations. And, finally, we use the FEAT~\footnote{https://github.com/lacava/feat} implementation written by the authors in C++ with a Python wrapper and using the Shogun C++ ML library~\citep{shogun}.
SymTree~\footnote{https://github.com/folivetti/ITSR} algorithm was implemented in Python with the use of the Scikit-Learn library to generate the affine combination of the IT terms and ITEA was developed in Haskell using GHC $8.6.5$~\footnote{available after review}.

It is also worth mentioning the algorithms Fast Function Extractor~\cite{mcconaghy2011ffx}, Evolutionary Feature Synthesis~\cite{arnaldo2015building} and Multiple Regression Genetic Programming~\cite{arnaldo2014multiple} that, similarly to IT-based algorithms and FEAT, search for an affine combination of subexpressions. Those approaches were extensively compared against SymTree in~\cite{de2018greedy} and FEAT in~\cite{LaCava:2019:SVO:3321707.3321776} and since they were ranked lower, they were left out from this comparison for the sake of brevity and to include different flavors of Genetic Programming.

\begin{table}[t!]
\centering
\caption{Average rank of each mutation operators subset (upper half) and the influence of a crossover operator (bottom half).}
\begin{tabular}{@{}lc@{}}
\toprule
Mutation set & Avg. Rank \\
\midrule
Add and Drop terms & $4.125$ \\
Change Interaction & $5.375$ \\
Pos. and Neg. Interactions & $5.5$ \\
Change Transformation &  $7.0$ \\

Add and Drop terms, Change Interaction & $2.625$ \\
Add and Drop terms, Change Interaction, Pos. and Neg. Interactions & $\mathbf{1.375}$ \\
All operators & $2.5$ \\
\bottomrule
\end{tabular}
\label{tab:medianisomut}
\end{table}

\subsubsection{Parameters setup}

The optimal hyper-parameters were chosen through a Grid Search procedure using a $5$-fold cross validation process on the training set. The values for the main parameters considered in this procedure for each algorithm is reported in Table~\ref{tab:params}. For reproducibility purpose, the full configurations is provided at our repository~\footnote{available after review}.

The function set used for the Symbolic Regression algorithms was $\{add, mul, sub, pdiv, \sin, \cos, pow, plog, \sqrt{|.|}, \tanh, exp\}$. For SymTree and ITEA, we only used the univariate functions from this set as required for the transformation function. Also, as an additional experiment, we have tested FEAT  with the the recommended parameters set and the additional functions $\{square, cube, |.|, logit, gauss, relu, \land, \lor, \neg, \oplus, =, <, >, \leq, \geq \}$ from~\cite{LaCava:2019:SVO:3321707.3321776}, tested with $100$ different data sets. This version will be denoted as FEAT-full in the following.

\section{Experimental Results}
\label{sec:results}

In this section we report the obtained results from the experiments described in the previous section and discuss how ITEA performs with respect to the literature.

\subsection{Importance of the Mutation Operators}

In Table~\ref{tab:medianisomut}  we can see the average rank for each of the considered subsets of mutation operators. The results point out that every mutation operator, except for the \emph{Change Transformation}, contributes to the performance of the algorithm. For the next set of experiments, we will use this optimal set.

\begin{table}[t!]
\caption{Mean RMSE and Std. Dev. of the training set for the data sets with dimension lower than $10$. The LR row shows the best results from the linear regression approaches.  The best result of each section  with multiple algorithms of the test set is highlighted in bold.}
\label{tab:resultslow}
\centering
\small
\begin{tabular}{l>{$}r<{$}>{$}r<{$}>{$}r<{$}>{$}r<{$}>{$}r<{$}>{$}r<{$}}
\toprule
\textbf{Alg.} & \text{Airfoil} & \text{Concrete} & \text{Cooling} & \text{Heating} & \text{Yacht} \\
\midrule
 & \multicolumn{5}{c}{Training Set} \\
 \midrule
ITEA & 2.27 \pm 0.06 & 5.81 \pm 0.11 & 1.44 \pm 0.05 & 0.46 \pm 0.02 & 0.49 \pm 0.06 \\
SymTree & 1.50 \pm 0.08 &   3.67 \pm 0.14 &  1.12 \pm 0.05 &  0.48 \pm 0.02  &  1.21 \pm 0.35 \\
FEAT-full  &   3.06 \pm 0.33 &   6.00 \pm 0.29 &  1.60 \pm 0.04 &  0.47 \pm 0.01 &  0.54 \pm 0.13 \\
FEAT &  3.48 \pm 0.46  & 6.63 \pm 0.44   & 1.72 \pm  0.13  & 0.93 \pm 0.41  & 1.73 \pm 0.63  \\
\midrule
dCGP          & 7.56\pm1.33    & 19.45\pm2.42    & 12.89\pm2.35   & 13.74\pm2.02   & 17.09\pm1.76 \\
GP            & 10.28\pm7.26   & 13.64\pm1.89    & 4.44\pm0.91    & 4.67\pm0.92    & 6.38\pm2.45  \\
GSGP          & 10.60\pm4.62   & 7.24\pm0.74     & 2.07\pm0.43    & 1.50\pm0.45    & 8.81\pm1.11  \\
\midrule
Forest        & 1.80\pm0.03    & 2.97\pm0.07     & 1.22\pm0.03    & 0.40\pm0.01    & 0.04\pm0.02  \\
Tree          & 2.04\pm0.08    & 2.90\pm0.16     & 1.68\pm0.07    & 0.34\pm0.03    & 0.33\pm0.41  \\
kNN           & 5.90\pm0.08    & 8.27\pm0.09     & 1.97\pm0.04    & 2.37\pm0.06    & 8.75\pm0.68  \\
\midrule
LR  & 4.82\pm0.04    & 10.47\pm0.12    & 3.22\pm0.02    & 2.96\pm0.05    & 8.94\pm0.15 \\
\midrule

 & \multicolumn{5}{c}{Test Set} \\
\midrule
ITEA &  2.45 \pm 0.21 & 6.33 \pm 0.47 & \mathbf{1.53 \pm 0.13} & \mathbf{0.49 \pm 0.04} & \mathbf{0.75 \pm 0.58} \\
SymTree &  \mathbf{2.12 \pm 0.20} &   \mathbf{5.56 \pm 0.38} &  1.61 \pm 0.17 &  0.64 \pm 0.04 &  1.61 \pm 0.22 \\
FEAT-full  &   3.18 \pm 0.31 &   6.35 \pm 0.28 &  1.63 \pm 0.12 &  \mathbf{0.49 \pm 0.05} &  0.83 \pm 0.22 \\
FEAT &  3.55 \pm 0.44  & 7.06 \pm 1.09   & 1.75 \pm 0.16  & 0.96 \pm 0.41  & 1.83 \pm 0.65  \\
\midrule
dCGP          & \mathbf{7.55\pm1.38}    & 19.50\pm2.27    & 12.84\pm2.25   & 13.67\pm1.88   & 17.01\pm2.20  \\
GP            & 10.31\pm7.21   & 13.65\pm2.09    & 4.46\pm0.90    & 4.67\pm0.97    & \mathbf{6.68\pm2.94}  \\
GSGP          & 10.60\pm4.35   & \mathbf{7.86\pm1.10}     & \mathbf{2.13\pm0.42}    & \mathbf{1.57\pm0.46}    & 9.03\pm1.03  \\
\midrule
Forest        & \mathbf{2.37\pm0.07}    & \mathbf{5.25\pm0.32}     & \mathbf{1.79\pm0.12}    & \mathbf{0.57\pm0.10}    & \mathbf{1.07\pm0.31}  \\
Tree          & 2.93\pm0.10    & 6.50\pm0.50     & 1.85\pm0.14    & \mathbf{0.57\pm0.07}    & 1.44\pm0.26  \\
kNN           & 5.86\pm0.26    & 7.85\pm0.73     & 1.92\pm0.13    & 2.32\pm0.42    & 7.56\pm1.31  \\
\midrule
LR & 4.82\pm0.17    & 10.44\pm0.49    & 3.21\pm0.11    & 2.94\pm0.19    & 8.93\pm0.71  \\
\bottomrule
\end{tabular}
\end{table}

\subsection{Comparison with other algorithms}

The mean and standard deviation of the RMSE obtained by each algorithm are reported in Tables~\ref{tab:resultslow}~and~\ref{tab:resultshigh}. These tables are divided into four sections: the top section contains the  Symbolic Regression algorithms that use affine combinations of subexpressions, the second section contains the GP variations, the third section has the nonlinear models, and the final section shows the best results obtained by the linear models.

Comparing with the linear models, ITEA had a smaller extrapolation error in every data set with the exception of \emph{Tecator} and \emph{White Wine}. When contrasting with the non-linear regression group, ITEA found better results than \emph{kNN} and \emph{Regression Tree} in $7$ of the data sets, and surpassed \emph{Random Forest} in $4$ data sets. Notice that \emph{Random Forest} found the best results in $5$ data sets when comparing to any other approach.

\begin{table}[t!]
\caption{Mean RMSE and Std. Dev. of the training set for the data sets with dimension greater or equal than $10$. The LR row shows the best results from the linear regression approaches. The best result of each section  with multiple algorithms of the test set is highlighted in bold.}
\label{tab:resultshigh}
\centering
\small
\begin{tabular}{l>{$}r<{$}>{$}r<{$}>{$}r<{$}>{$}r<{$}>{$}r<{$}>{$}r<{$}}
\toprule
\textbf{Alg.} & \text{Geo}   & \text{Tower}   & \text{Tecator} & \text{Red}  & \text{White} \\
\midrule
 & \multicolumn{5}{c}{Training Set} \\
 \midrule 
ITEA & 32.94 \pm 2.21 & 22.2 \pm 0.57 & 0.7 \pm 0.09 & 0.61 \pm 0.01 & 0.70 \pm 0.00 \\
SymTree & 31.95 \pm 0.90 &  12.61 \pm 0.36 & 0.42 \pm 0.04 & 0.44 \pm 0.01 & 0.49 \pm 0.01 \\
FEAT-full & 41.96 \pm 4.35 & 20.46 \pm 1.92 &  1.02 \pm 0.54 & 0.63 \pm 0.01 &  0.73 \pm 0.01 \\
FEAT &  41.87 \pm 1.44  &  25.59 \pm 4.33  & 0.83 \pm 0.30  & 0.63 \pm 0.01  & 0.72 \pm 0.01  \\
\midrule
dCGP          & 55.60\pm3.41 & 106.34\pm24.03 & 4.10\pm0.35    & 1.14\pm0.97 & 1.14\pm0.84  \\
GP            & 49.66\pm2.58 & 68.34\pm5.87   & 3.00\pm0.63    & 0.79\pm0.06 & 0.87\pm0.05  \\
GSGP          & 42.03\pm0.96 & 63.84\pm4.70   & 1.04\pm0.46    & 0.62\pm0.01 & 0.73\pm0.02  \\
\midrule
Forest        & 11.26\pm0.76 & 12.97\pm0.42   & 0.63\pm0.02    & 0.34\pm0.00 & 0.53\pm0.00  \\
Tree          & 19.73\pm3.50 & 13.55\pm0.52   & 0.66\pm0.21    & 0.60\pm0.01 & 0.70\pm0.01  \\
kNN           & 31.51\pm0.52 & 15.11\pm0.14   & 1.39\pm0.04    & 0.68\pm0.01 & 0.71\pm0.01  \\
\midrule
LR & 43.70\pm0.18 & 30.31\pm1.05   & 0.80\pm0.03    & 0.65\pm0.01 & 0.75\pm0.00   \\
\midrule

 & \multicolumn{5}{c}{Test Set} \\
\midrule
ITEA & \mathbf{35.65 \pm 2.69} & \mathbf{23.03 \pm 1.56} & 1.2 \pm 0.37 & \mathbf{0.64 \pm 0.03} & 0.77 \pm 0.19  \\
SymTree & 43.53 \pm 1.49 &    23.67 \pm 11.33 & \mathbf{0.71 \pm 0.16} & \mathbf{0.64 \pm 0.03} & 0.80 \pm 0.18  \\
FEAT-full & 45.11 \pm 9.02 & 25.00 \pm 9.85 & 2.02 \pm 4.93 & 0.65 \pm 0.03 &  \mathbf{0.76 \pm 0.08} \\
FEAT &  45.67 \pm 9.12  &  32.82 \pm 19.95  & 1.03 \pm 0.32  & 0.65 \pm 0.03  & 5.19 \pm 15.42  \\
\midrule
dCGP          & 63.75\pm46.90 & 106.93\pm24.39 & 4.10\pm0.43    & 1.14\pm0.97 & 1.14\pm0.83  \\
GP            & 49.72\pm2.94  & 68.61\pm6.59   & 3.05\pm0.70    & 0.79\pm0.07 & 0.88\pm0.05  \\
GSGP          & \mathbf{44.35\pm1.51}  & \mathbf{64.00\pm5.15}   & \mathbf{1.41\pm0.56}    & \mathbf{0.64\pm0.03} & \mathbf{0.74\pm0.02}  \\
\midrule
Forest        & \mathbf{25.78\pm3.65}  & 17.69\pm1.27   & \mathbf{1.25\pm0.12}    & \mathbf{0.58\pm0.04} & \mathbf{0.66\pm0.01}  \\
Tree          & 31.92\pm7.42  & 20.09\pm1.02   & 1.68\pm0.15    & 0.69\pm0.06 & 0.75\pm0.02  \\
kNN           & 31.17\pm0.87  & \mathbf{14.81\pm0.85 }  & 1.29\pm0.09    & 0.66\pm0.05 & 0.69\pm0.02  \\
\midrule
LR & 42.95\pm1.26  & 30.56\pm3.46   & 0.74\pm0.06    & 0.65\pm0.03 & 0.75\pm0.01  \\
\bottomrule
\end{tabular}
\end{table}

Regarding the second block, composed of GP variations, ITEA found better results in every data set, with the exception of Wine White, where \emph{GSGP} found a slightly better expression.
Finally, among the first group, ITEA found expressions with equal or smaller extrapolation errors in $6$ data sets.  SymTree in $4$ data sets and \emph{FEAT-full} in only $2$. FEAT did not perform well comparatively to this group, but every algorithm in this group was significantly better than the \emph{GP} and the \emph{Linear} groups. 
Specifically for the data sets with more than $100$ features, we can see that for the Geographical data set, ITEA was the only algorithm among the Symbolic Regression approaches to find a better approximation than the linear models. In \emph{Tecator}, SymTree was the only algorithm with better average RMSE when compared to the linear models. This happened due to its constructive nature that starts with a linear regression and adds new features that improves the current solution, notice though that the best model found by ITEA was on par with SymTree (see Table~\ref{tab:besttest}).
The statistical significance of these results is reported in Table~\ref{tab:pvals}.
In Table~\ref{tab:besttest} we report the best obtained result for \emph{ITEA, SymTree, FEAT-full} algorithms. We can see that both ITEA and SymTree found the best expression in $5$ data sets and \emph{FEAT-full} only in $3$. Even when ITEA did not find the best, it almost always found the second best expression.

\begin{table}[t!]
\caption{RMSE of the best results found by FEAT-full, ITEA and SymTree for the test set. The best results for each row is highlighted in bold and the second best in underline.}
\label{tab:besttest}
\centering
\begin{tabular}{l>{$}r<{$}>{$}r<{$}>{$}r<{$}}
\toprule
Algorithm     &   \text{FEAT-full} &   \text{ITEA} & \text{SymTree} \\
Dataset       &        &        &         \\
\midrule
airfoil       &   2.65 &   \underline{2.11} &    \mathbf{1.89} \\
concrete      &   5.64 &   \underline{5.49} &    \mathbf{5.10} \\
energyCooling &   1.47 &   \mathbf{1.33} &    \underline{1.41} \\
energyHeating &   \mathbf{0.41} &   \mathbf{0.41} &    \underline{0.59} \\
Geographical  & \mathbf{19.88} & \underline{28.67} & 42.10 \\
tecator       & 0.66 & \underline{0.57} & \mathbf{0.56} \\
towerData     &  \underline{18.63} &  20.52 &   \mathbf{17.69} \\
wineRed       &   \mathbf{0.60} &   \mathbf{0.60} & \underline{0.62} \\
wineWhite     &   \underline{0.71} &   \mathbf{0.70} &     \mathbf{0.70} \\
yacht         &   \underline{0.48} &   \mathbf{0.46} &    1.26 \\
\bottomrule
\end{tabular}
\end{table}

\begin{figure}[t!]
\centering
\begin{subfigure}[b]{0.49\textwidth}
\includegraphics[trim=0.8cm 1.7cm 3cm 3cm,clip,width=\textwidth]{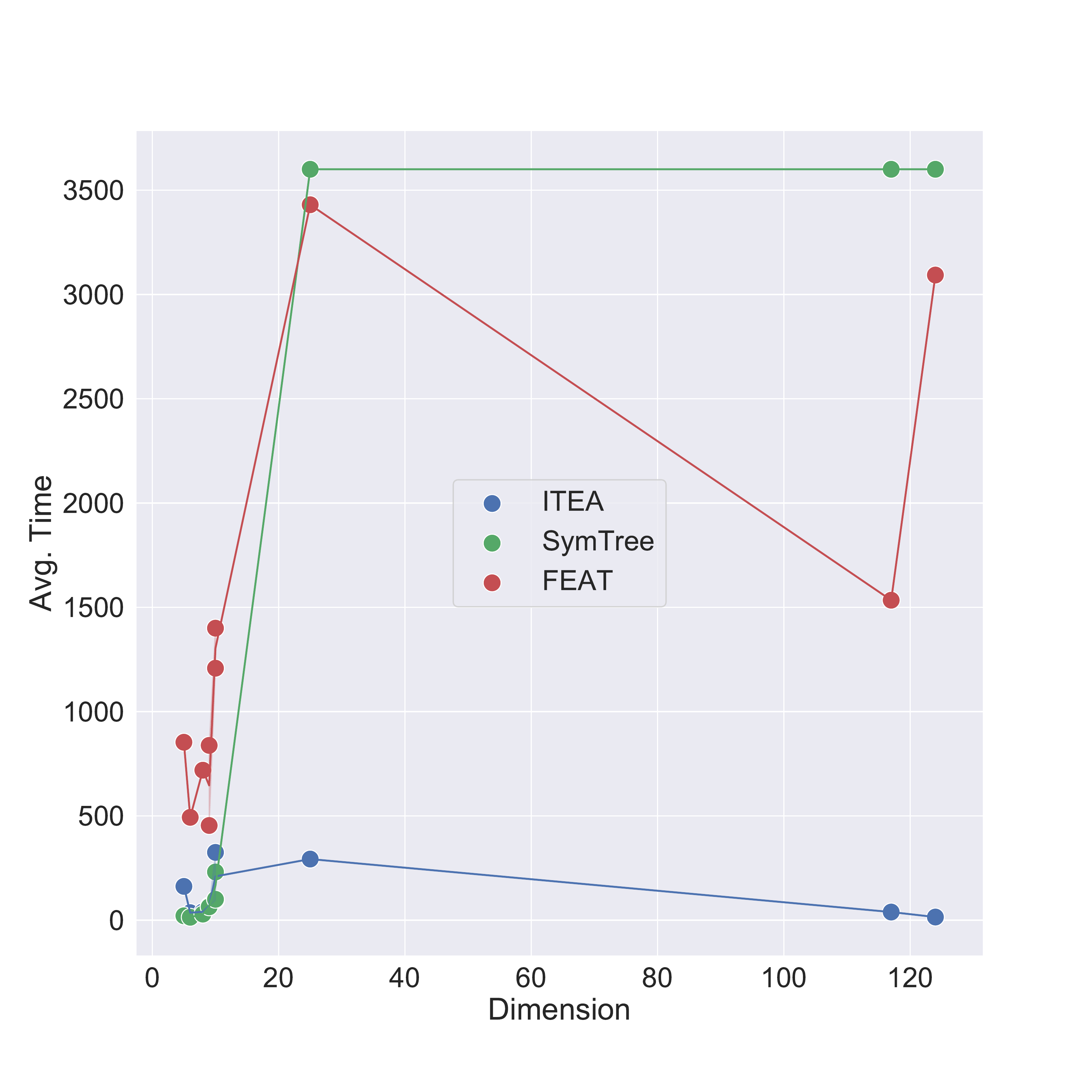}
\caption{}
\label{fig:time}
\end{subfigure}
\begin{subfigure}[b]{0.49\textwidth}
\includegraphics[trim=0.1cm 1.7cm 3cm 3cm,clip,width=\textwidth]{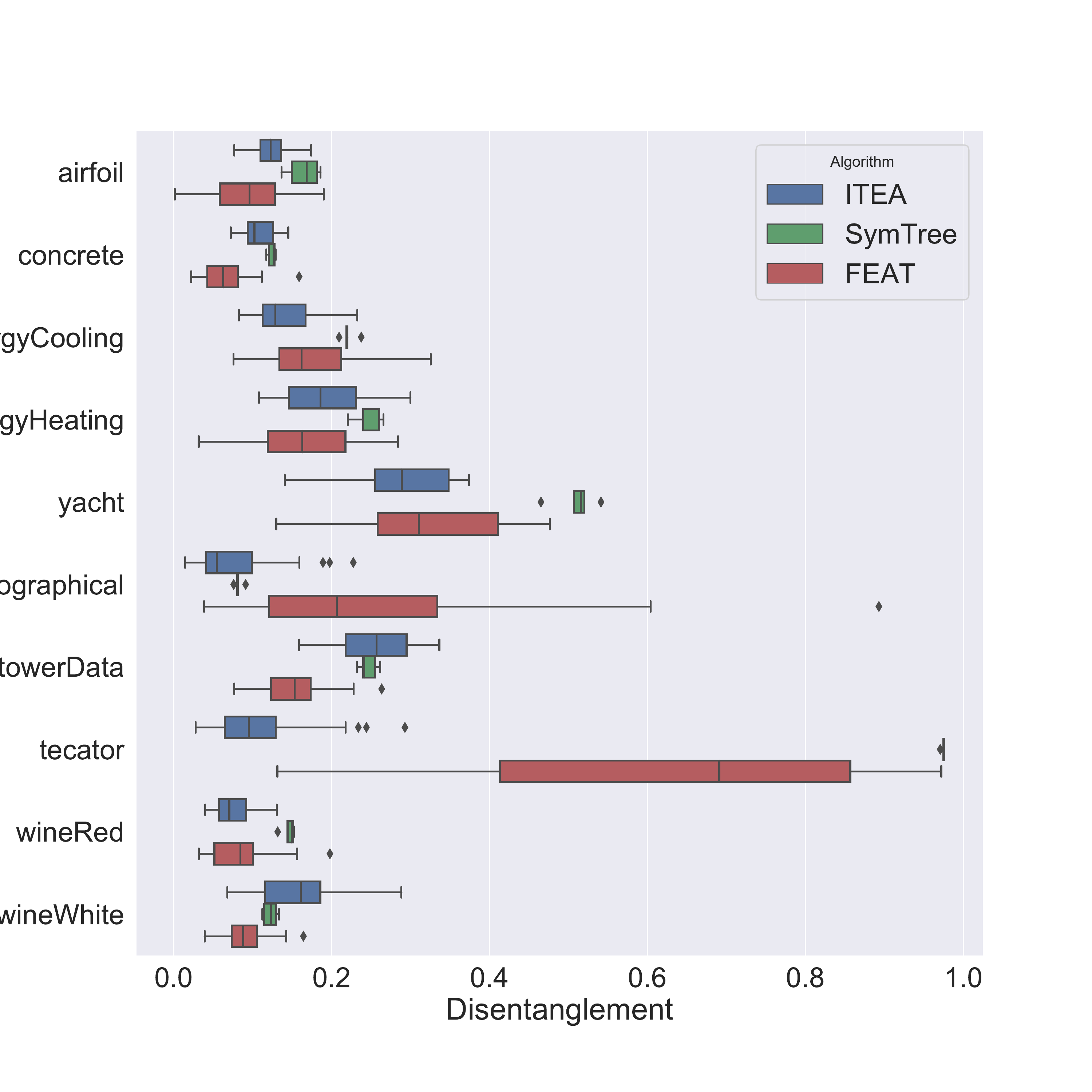}
\caption{}
\label{fig:disent}
\end{subfigure}
\caption{(a) Average execution time w.r.t. data set dimension and (b) Average disentanglement of the expressions obtained by ITEA, SymTree and FEAT for each data set.}
\label{fig:timedisen}
\end{figure}

In Fig.~\ref{fig:time}, we can see the average execution time of \emph{ITEA, SymTree} and FEAT w.r.t. the data set dimension. While we cannot make a direct comparison of these values due to differences in implementation, we can verify how they scale with the problem dimension. As we can see from this plot, ITEA execution time was the least affected by the data set dimension. SymTree and FEAT presented an exponential growth of execution time limited by the maximum value allowed.

Finally, in Fig.~\ref{fig:disent}, we compare the disentanglement of the Symbolic Regression algorithms with affine combination. The expressions generated by these algorithms can be seen as a transformed representation of the original features set. The disentanglement measures the multicollinearity of these new features. Following the work in~\cite{cava2018learning}, this was calculated as the mean of the absolute value of the Pearson correlation between each pair of features. In this plot, we can see that all three algorithms behaved similarly across the data sets with a disentanglement value between $0.0$ and $0.2$. The few exceptions can be observed in \emph{Yacht}, where ITEA and FEAT was both within the range $[0.2, 0.4]$ and SymTree within $[0.4, 0.5]$, and \emph{Tecator}, where FEAT disentanglement ranged within $[0.4, 0.9]$ and SymTree maintained a value close to $1.0$.






\subsection{Measuring the Importance of the Predictors}
\label{sec:importance}

Another important aspect of a regression model is to understand the role of the predictors in the studied system. One way to understand this role is by measuring the importance of the original variables in the prediction process.
With a linear model, the importance of each variable is given by the adjusted coefficients ($\beta$ in Eq.~\ref{eq:linreg}). A value of $\beta_i$ corresponding to the $i$-th predictor says that, for every unit we add to $x_i$, the target value is increased by $\beta_i$.

In the non-linear models generated by Symbolic Regression, this relationship is not as straightforward. The difficulty here is that the importance of each predictor may depend on the other predictors or may vary depending on its own value. One way to deal with this problem is to apply model agnostic approaches~\cite{ribeiro2016should,lundberg2017unified}. The main benefit of these approaches is that they work with any regression and classification model, even those considered to be black-boxes. But, they are just approximations by a local linear model focusing on a point of interest.

Given that Symbolic Regression returns an analytical solution as a model, we can return the importance of each predictor not as a value, but as an analytical expression. Notice that the importance of a predictor $x_i$ is given by the partial derivative (if one exists) $\dfrac{\partial f}{\partial x_i}$. For example, the importance of a linear model is reduced to the coefficients when calculating the gradient vector.
A particular advantage of the IT representation, is that we can easily automate this process since the returned expression follows a well structured closed form. The derivative of an IT expressions w.r.t. any given variable $x_j$ is given by:

\begin{equation*}
\frac{\partial IT(x)}{\partial x_j} = w_1 \cdot g'_1(x) + \ldots + w_n \cdot g'_n(x),
\end{equation*}

\noindent with

\begin{align*}
g'_i(x) &= t'_i(p_i(x)) \cdot p'_i(x) \\
p'_i(x) &= k_j\frac{p_i(x)}{x_j},
\end{align*}

\noindent so, by providing the derivatives of the set of transformation functions, we can readily return the corresponding gradient vector.

On the other hand, with standard Symbolic Regression approaches that imposes no restrictions to the function form, we cannot have a closed form approach to find the derivatives and thus we must resort to Symbolic Differentiation or Automatic Differentiation~\cite{baydin2017automatic}. Symbolic Differentiation often requires a high computational cost and may return a complex expression as a result. Automatic Differentiation often requires a modification of the expression in the form of a source code and, as a result, it will only return a black-box function that computes the derivative at a given point. We should notice, though, that many of the expressions generated by FEAT during the experiments could be rewritten as an IT expression. So, with some exceptions, we could apply the same treatment with the expressions found by FEAT without much effort. But, in some situations, FEAT generated expressions with a chaining of nonlinear functions or the use of non-standard operations such as $if-then-else$, which made the process of automating the calculation of derivatives harder to accomplish.

To illustrate the procedure of evaluating the importance of a predictor using an IT expression, we generated expressions with ITEA limited to $5$ terms for the \emph{Concrete} data set. This data set describes the concrete compressive strength given the properties described in Table~\ref{tab:heatfeats}~\cite{yeh1998modeling}.
The generated expression and the partial derivatives are reported in Table~\ref{tab:exprs}, except for $x_2$ that is not a part of the expression.

Given these partial derivatives, we can visualize the expected importance of each variable within their domains by calculating the Marginal Effect for each variable. The Marginal Effect of a variable $x_i$ replaces the values of each segment of the expression that does not involve this particular variable with the expected value. So, the partial derivatives from Table~\ref{tab:exprs} are all turned into univariate functions.

\begin{table}[t!]
    \centering
    \caption{Predictors of the Concrete data set~\cite{yeh1998modeling}.}
    \begin{tabular}{cc}
        \toprule 
        \textbf{Predictor} & \textbf{Description} \\
        \midrule
        $x_0$ & kilograms of cement in a $m^3$ mixture \\
        $x_1$ & kilograms of blast furnace slag in a $m^3$ mixture \\
        $x_2$ & kilograms of fly ash in a $m^3$ mixture \\
        $x_3$ & kilograms of water in a $m^3$ mixture \\
        $x_4$ & kilograms of superplasticizer in a $m^3$ mixture \\
        $x_5$ & kilograms of coarse aggregate in a $m^3$ mixture \\
        $x_6$ & kilograms of fine aggregate in a $m^3$ mixture \\
        $x_7$ & age \\
        \bottomrule
    \end{tabular}
    \label{tab:heatfeats}
\end{table}

\begin{figure}[t!]
\centering
\begin{subfigure}[b]{0.3\textwidth}
\includegraphics[width=\textwidth]{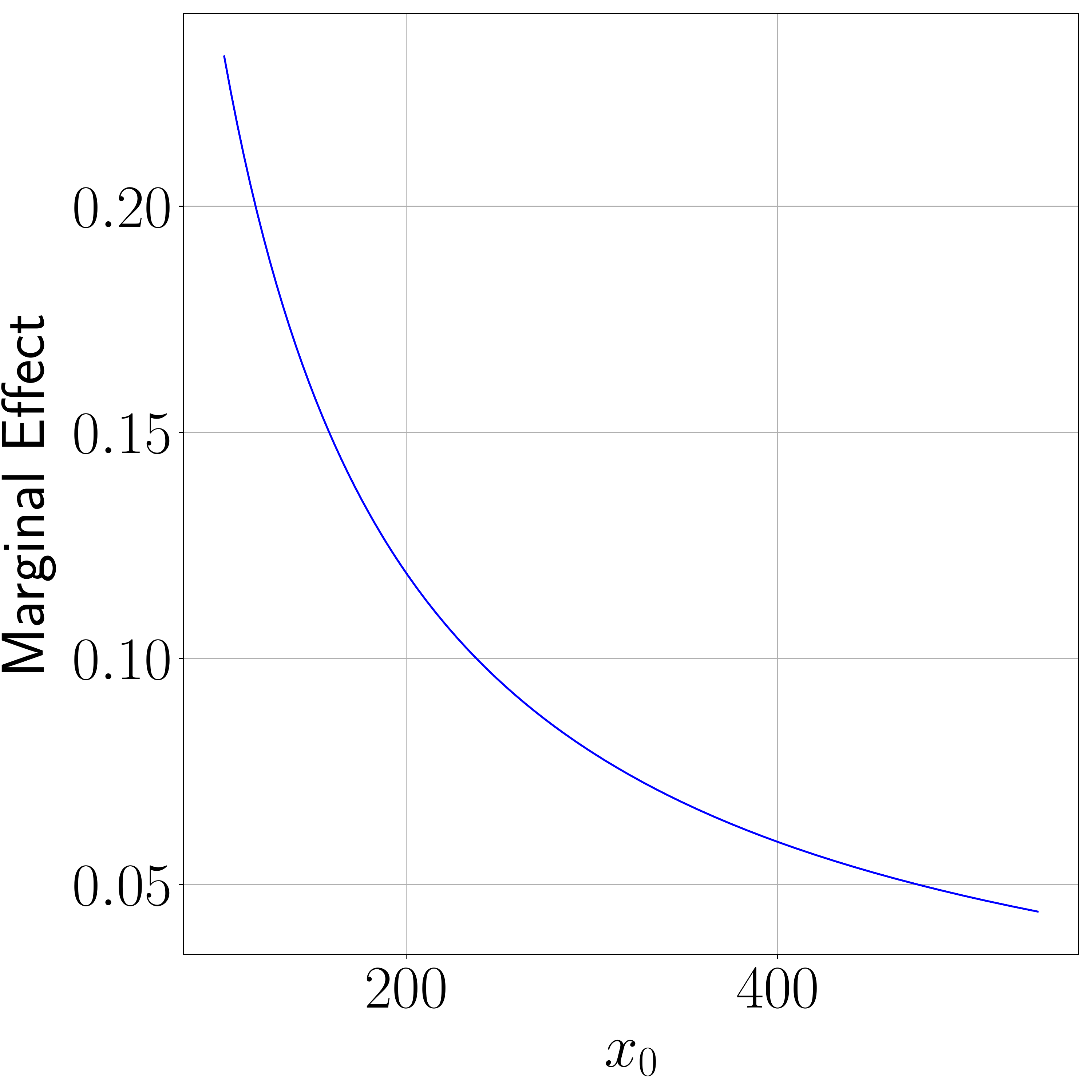}
\caption{}
\label{fig:mex0}
\end{subfigure}
\vspace{1em}
\begin{subfigure}[b]{0.3\textwidth}
\includegraphics[width=\textwidth]{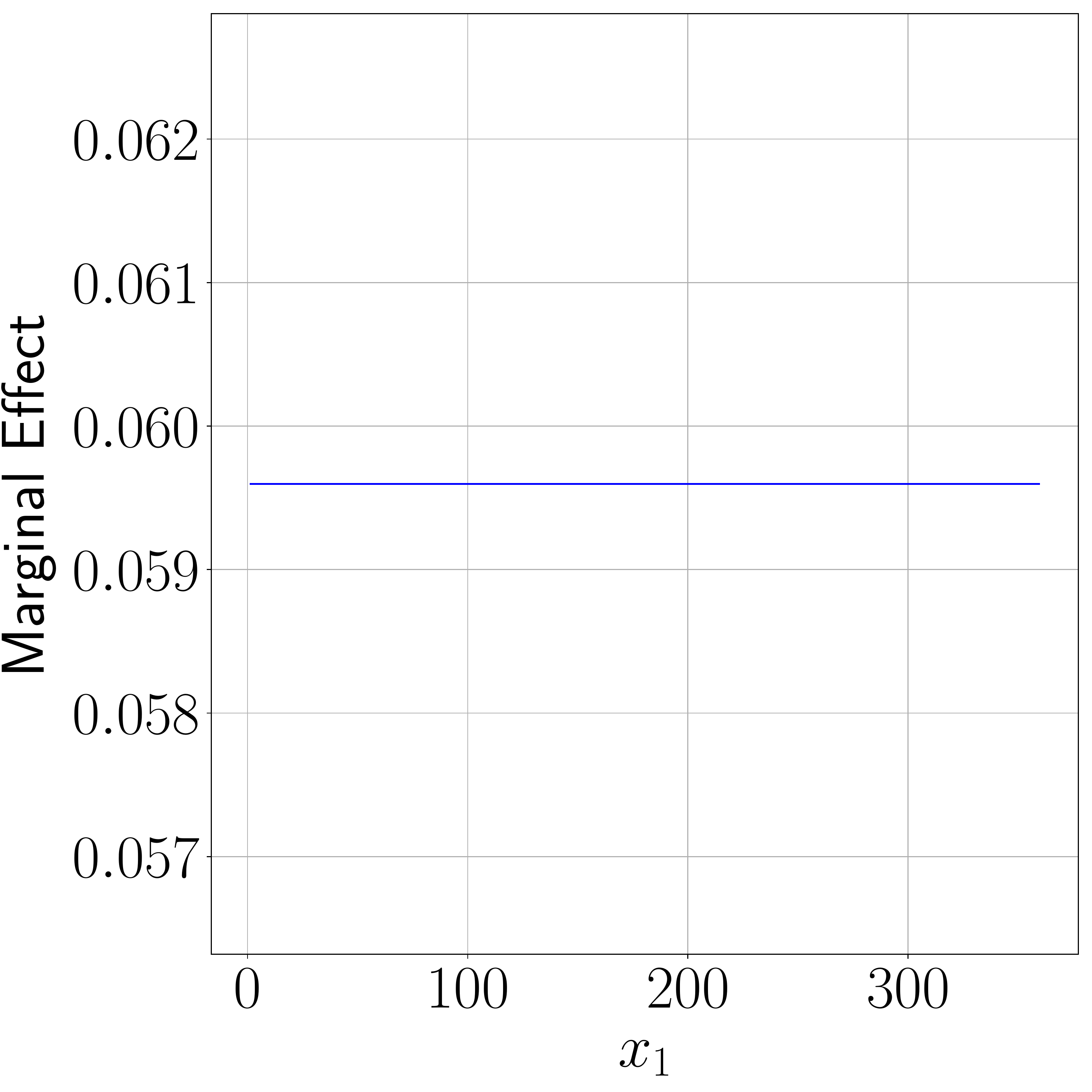}
\caption{}
\label{fig:mex1}
\end{subfigure}
\vspace{1em}
\begin{subfigure}[b]{0.3\textwidth}
\includegraphics[width=\textwidth]{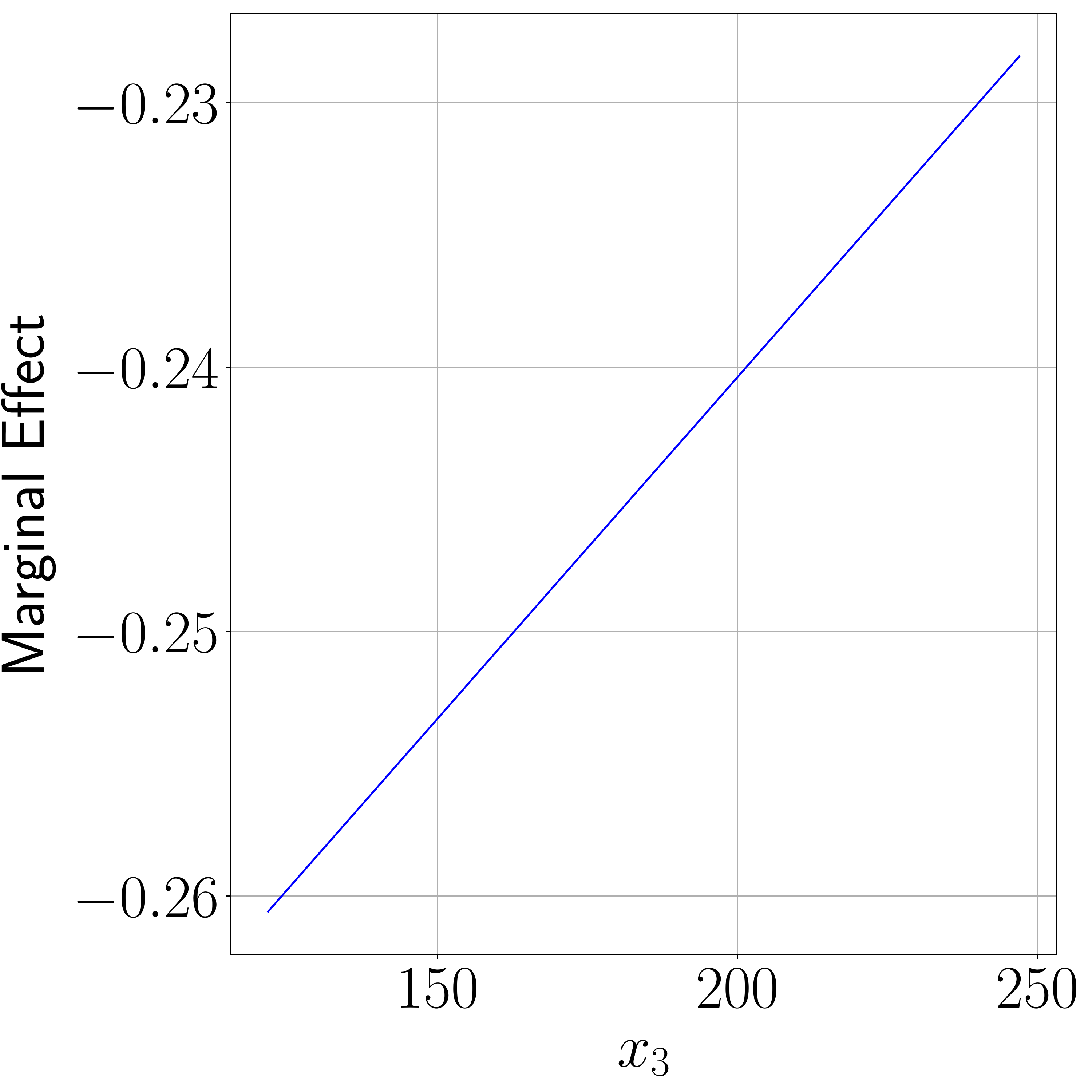}
\caption{}
\label{fig:mex3}
\end{subfigure}

\begin{subfigure}[b]{0.3\textwidth}
\includegraphics[width=\textwidth]{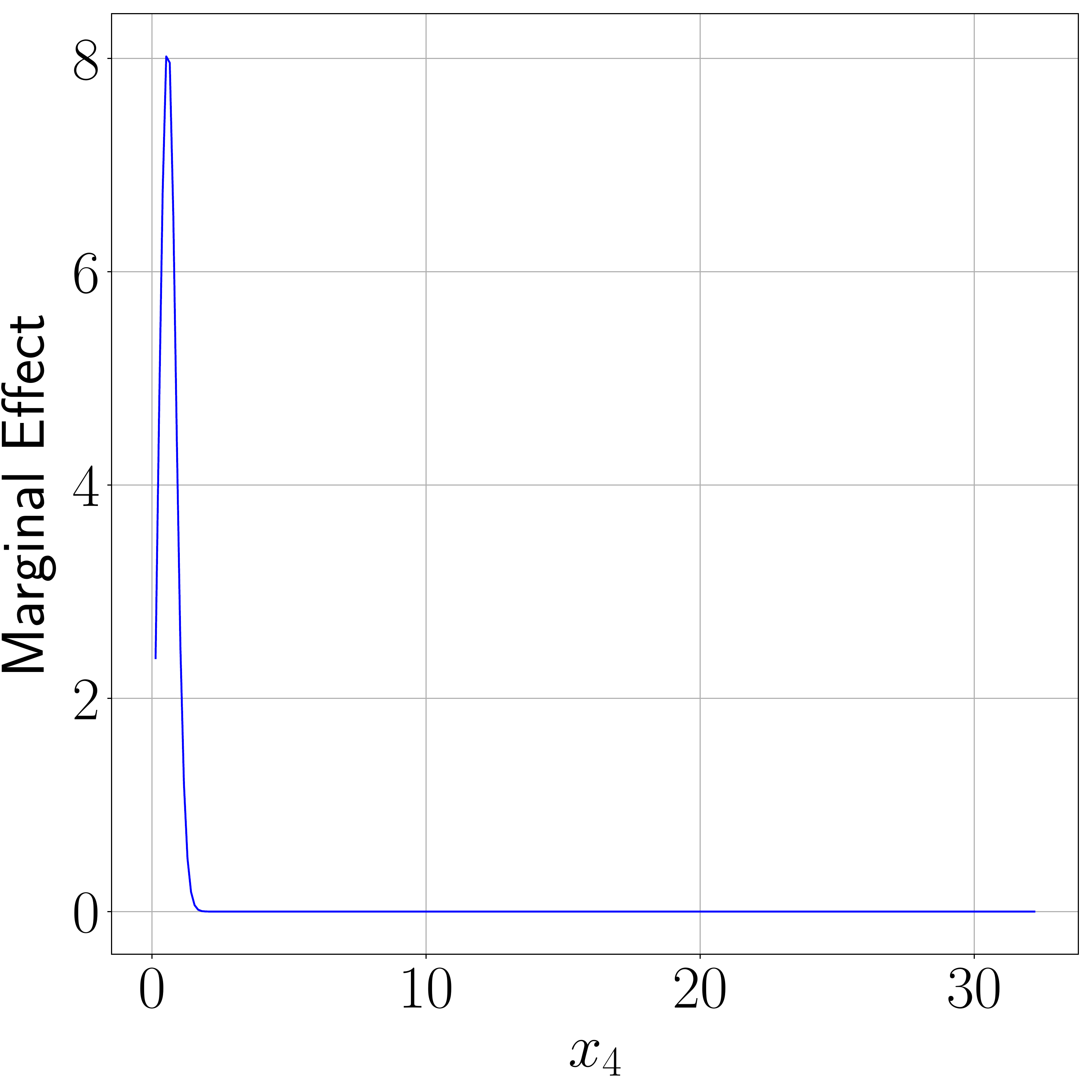}
\caption{}
\label{fig:mex4}
\end{subfigure}
\vspace{1em}
\begin{subfigure}[b]{0.3\textwidth}
\includegraphics[width=\textwidth]{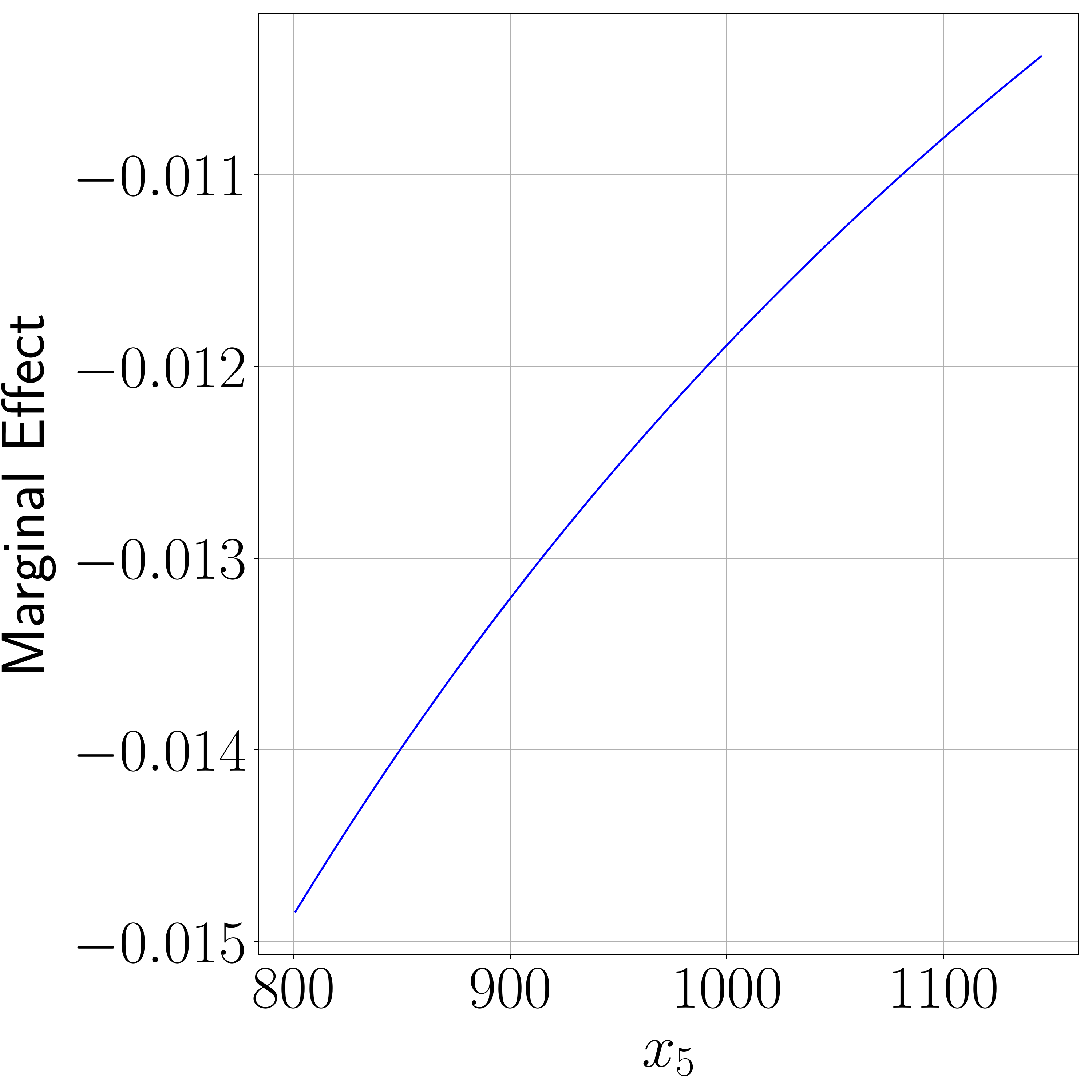}
\caption{}
\label{fig:mex5}
\end{subfigure}
\vspace{1em}
\begin{subfigure}[b]{0.3\textwidth}
\includegraphics[width=\textwidth]{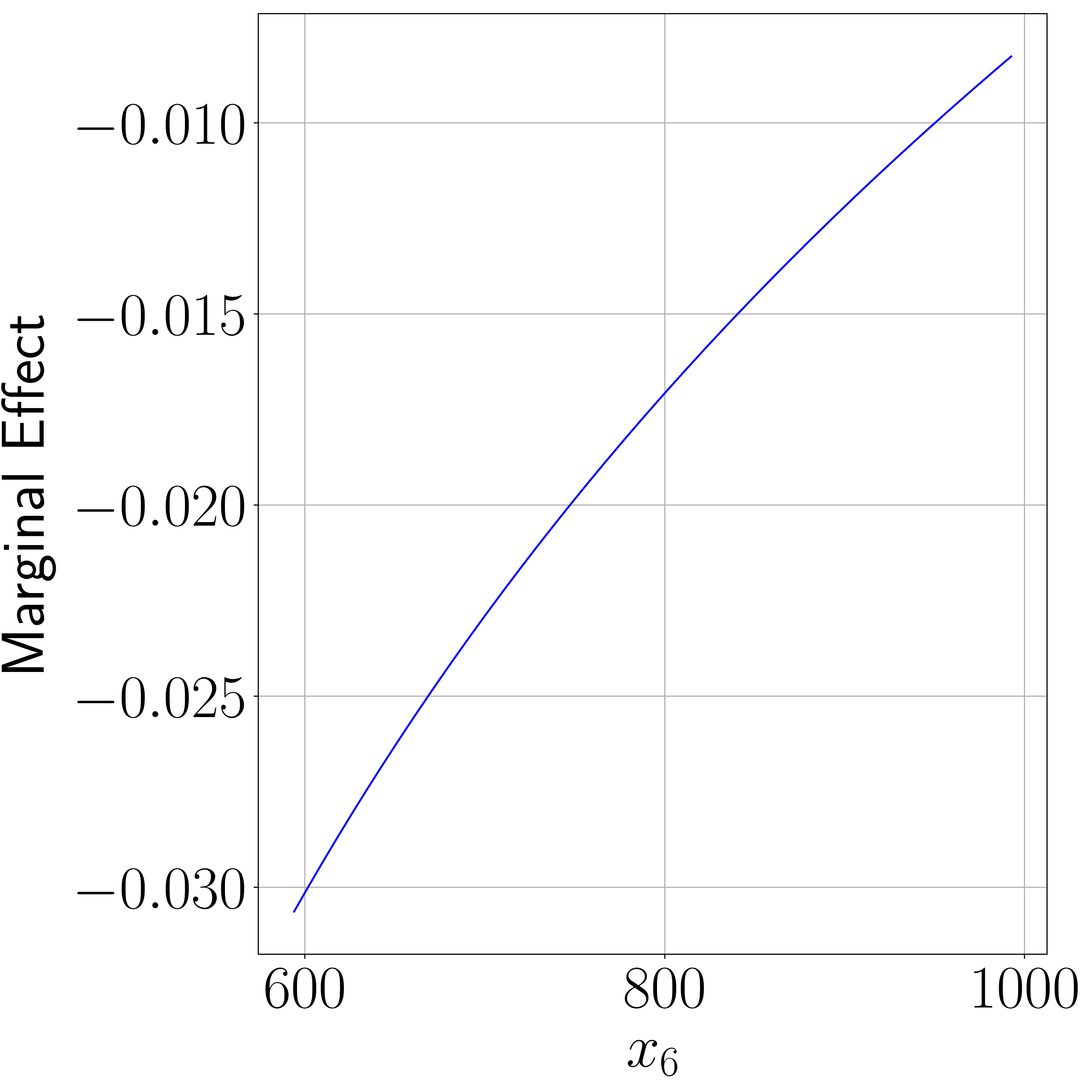}
\caption{}
\label{fig:mex6}
\end{subfigure}

\begin{subfigure}[b]{0.3\textwidth}
\includegraphics[width=\textwidth]{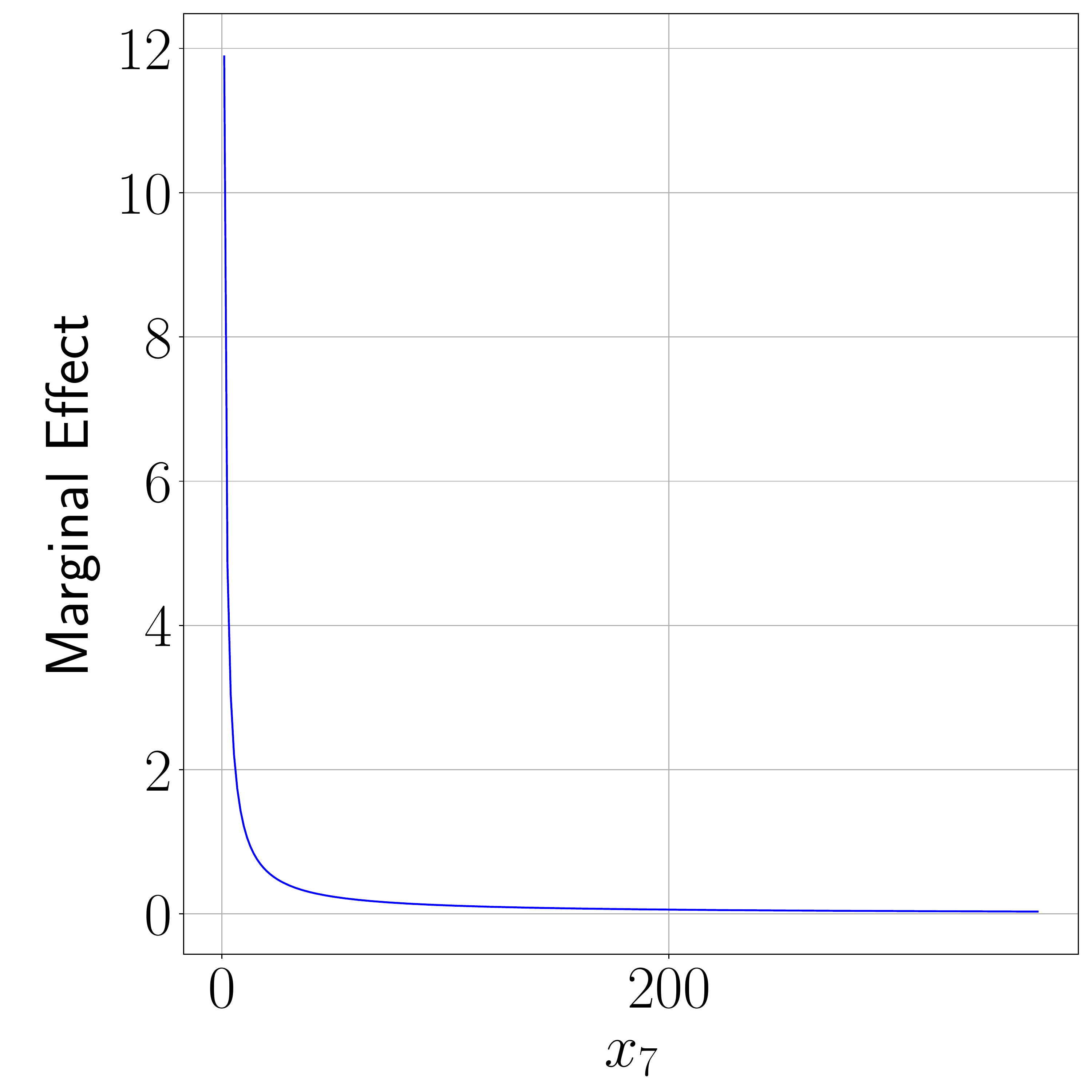}
\caption{}
\label{fig:mex7}
\end{subfigure}

\caption{Marginal effect of the original features of Concrete data set.}
\label{fig:me}
\end{figure}

In Fig.~\ref{fig:me} we can see the plot of the marginal effect of the variables for the above expression. From this plot we can see that some variables have a decreasing importance ($x_0, x_4, x_7$), some are increasing ($x_3, x_5, x_6$) and $x_1$ have a constant importance.
The decreasing importance, for example, states that the higher the value of this particular predictor, the smaller is the importance to the prediction. For example, \emph{age} is a decreasing factor and it has a large impact in the compressive strength up until 10 years. After this period, the strength is barely affected by this variable.

\begin{figure}[t!]
\centering
\begin{subfigure}[b]{0.48\textwidth}
\includegraphics[trim=2cm 1.7cm 3cm 3cm,clip,width=\textwidth]{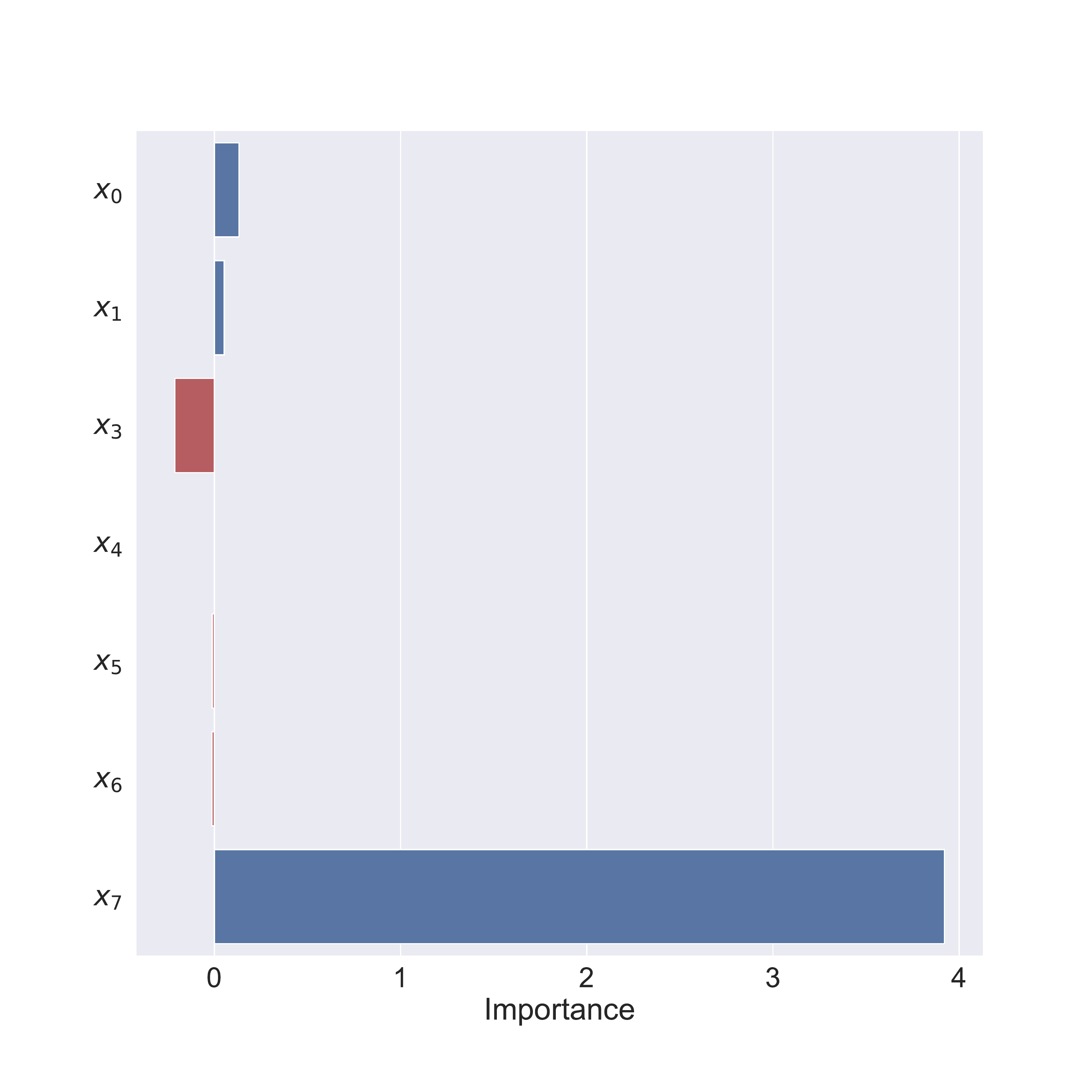}
\caption{}
\end{subfigure}
\begin{subfigure}[b]{0.48\textwidth}
\includegraphics[trim=2cm 1.7cm 3cm 3cm,clip,width=\textwidth]{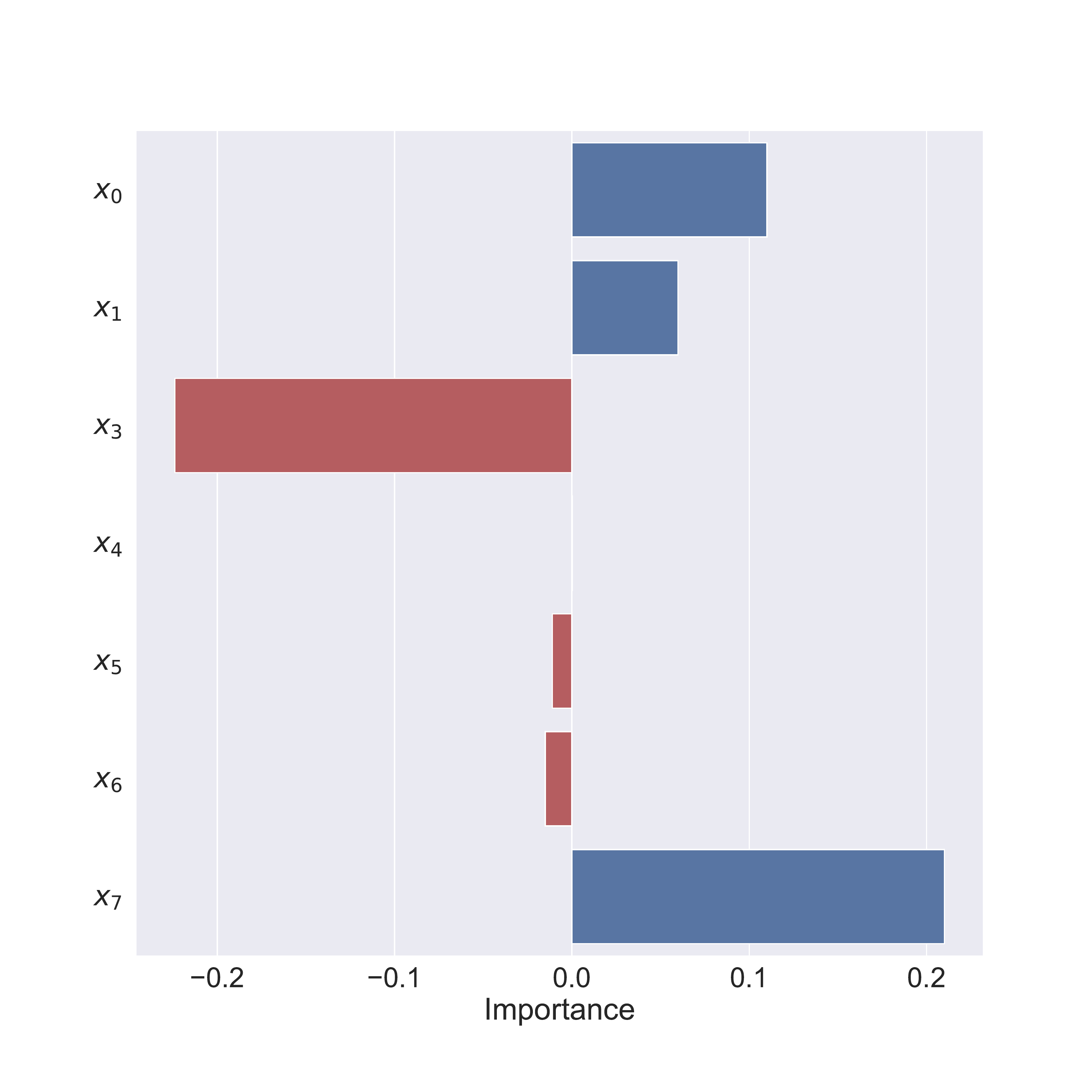}
\caption{}
\end{subfigure}
\caption{Importance of each feature when predicting the examples (a) $[178, 129.8, 118.6, 179.9, 3.6, 1007.3, 746.8, 3]$ and (b) $[213.8, 98.1, 24.5, 181.7, 6.7, 1066, 785.5, 56]$.}
\label{fig:barme}
\end{figure}

Another way to visualize the importance of each predictor is by plotting the local importance of a particular sample. The local importance of each variable is obtained by calculating the values of the gradient vector for the particular input data. In Fig.~\ref{fig:barme} we show the importance of each attribute for two examples: one with a small value of water and another with a higher value. We can see from these plots that the importance of age of the concrete is greatly reduced after a long time has passed and the other components become more decisive to the target variable.

\section{Conclusion}
\label{sec:conclusion}

The Interaction-Transformation is a recently proposed representation for the Symbolic Regression problem that reduces the search space to an additive model of non-linear terms w.r.t. the original variable space.
This representation was previously shown to be effective when compared to popular artificial benchmarks.
In this paper we have introduced a mutation-based evolutionary algorithm with the purpose of evolving IT-expressions, which we named ITEA. 
The mutation operator is composed of a random choice between six different mutation heuristics each of which changes one structural component of the expression by adding or removing a term, changing a nonlinear function, replacing part of the expression, or combining two terms through the interaction of variables.

For evaluation purposes, the algorithm was tested on $10$ different real-world data sets of different sizes and the obtained results were compared against those obtained by linear and nonlinear models and other symbolic regression approaches.
The results indicated that ITEA overperformed the other symbolic regression algorithms followed closely by SymTree, another algorithm using the IT representation. When compared to the linear and nonlinear models, ITEA competed with Random Forest algorithm and surpassed the results of the other models in most data sets. Regarding the generated terms of the expression, they presented a small pairwise collinearity as measured by the disentanglement.

Besides the quantitative performance, we have shown how the IT representation allows us to extract the importance of each original feature of the data set and explain a given prediction. Even though this can also be done with other Symbolic Regression approaches, the well structured representation of IT makes it easier to automate the process.

As for future research, we have different aspects of the algorithm that deserve a detailed investigation. For instance, sensitivity analysis of the parameters to verify whether there is an optimal set that best fit all scenarios or which ones should be adjusted w.r.t. the data set characteristic. 
Also, we intend to propose a set of crossover operators which is expected to improve the convergence towards an optimal solution. Following this proposal, we will investigate the potentials of a multi-population approach running on a distributed environment to deal with higher dimensional data sets. Finally, other bio-inspired approaches will be considered by adapting their core characteristics to the evolution of IT-expressions.

\section*{Acknowledgment}

This project is funded by Funda\c{c}\~{a}o de Amparo \`{a} Pesquisa do Estado de S\~{a}o Paulo (FAPESP), grant number 2018/14173-8. Also, the experiments made use of Intel\textregistered AI DevCloud, which Intel\textregistered provided free access.

\bibliographystyle{apalike}
\bibliography{citacoes}

\appendices 
\section{Partial derivatives of Concrete data set}

The following table contains the illustrative expression of concrete data set mentioned in Section~\ref{sec:importance}.

\begin{table}[!ht]
   \caption{Sample expression from ITEA for the Concrete data set limited to a maximum of $5$ terms and with coefficients rounded to the third decimal place. The RMSE value on the test set for this expression is $6.74$. We omitted the partial derivative of $x_2$ since it does not appear on the generated expression.}
 	\label{tab:exprs}
\begin{tabular}{lp{4.3in}}
\toprule
Function & Expression \\
\midrule
$f(\mathbf{x}) $ & $-0.003 \cdot \sqrt{|x_3 \cdot x_6 \cdot x_7|} - 11.762 \cdot \log(x_0^{-2} \cdot x_5 \cdot x_6^2 \cdot x_7^{-1}) + 5.790 \cdot \tanh(x_0^2 \cdot x_3 \cdot x_4^2 \cdot x_5^{-2} \cdot x_7^{-1}) - 0.288 \cdot x_3 + 0.000 \cdot (x_1 \cdot x_3^2 \cdot x_6^2) + 156.242$ \\
\midrule
$\dfrac{\partial f}{\partial x_0}$ & $23.525 \cdot \dfrac{log(\frac{x5 \cdot x6^{2}}{x0^{3} \cdot x7})}{\frac{x5 \cdot x6^{2}}{x0^{2} \cdot x7}}  + 11.579 \cdot \dfrac{tanh(\frac{x0 \cdot x3 \cdot x4^{2}}{x5^{2} \cdot x7})}{cosh(\frac{x0^{2} \cdot x3 \cdot x4^{2}}{x5^{2} \cdot x7})^2} $ \\
\midrule
$\dfrac{\partial f}{\partial x_1}$ & $0.000 \cdot x3^{2} \cdot x6^{2}$ \\
\midrule
$\dfrac{\partial f}{\partial x_3}$ & $-0.003 \cdot \dfrac{\sqrt{\left |x6 \cdot x7\right |}}{2 \cdot \left |x3 \cdot x6 \cdot x7\right | ^{\frac{3}{2}}} + 5.79 \cdot \dfrac{tanh(\frac{x0^{2} \cdot x4^{2}}{x5^{2} \cdot x7})}{cosh(\frac{x0^{2} \cdot x3 \cdot x4^{2}}{x5^{2} \cdot x7})^2}  + 0.0  \cdot x1 \cdot x3 \cdot x6^{2}$ \\
\midrule
$\dfrac{\partial f}{\partial x_4}$ & $11.579 \cdot \dfrac{tanh(\frac{x0^{2} \cdot x3 \cdot x4}{x5^{2} \cdot x7})}{cosh(\frac{x0^{2} \cdot x3 \cdot x4^{2}}{x5^{2} \cdot x7})^2}$ \\
\midrule
$\dfrac{\partial f}{\partial x_5}$ & $-11.762 \cdot \dfrac{log(\frac{x6^{2}}{x0^{2} \cdot x7})}{\frac{x5 \cdot x6^{2}}{x0^{2} \cdot x7}} + -11.579 \cdot \dfrac{tanh(\frac{x0^{2} \cdot x3 \cdot x4^{2}}{x5^{3} \cdot x7})}{cosh(\frac{x0^{2} \cdot x3 \cdot x4^{2}}{x5^{2} \cdot x7})^2} $ \\
\midrule
$\dfrac{\partial f}{\partial x_6}$ & $-0.003 \cdot \dfrac{\sqrt{\left |x3 \cdot x7\right |}}{2 \cdot \left |x3 \cdot x6 \cdot x7\right | ^{\frac{3}{2}}} + -23.525 \cdot \dfrac{log(\frac{x5 \cdot x6}{x0^{2} \cdot x7})}{\frac{x5 \cdot x6^{2}}{x0^{2} \cdot x7}} + 0.0  \cdot x1 \cdot x3^{2} \cdot x6$ \\
\midrule
$\dfrac{\partial f}{\partial x_7}$ & $-0.003 \cdot \dfrac{\sqrt{\left |x3 \cdot x6\right |}}{2 \cdot \left |x3 \cdot x6 \cdot x7\right | ^{\frac{3}{2}}} + 11.762 \cdot \dfrac{log(\frac{x5 \cdot x6^{2}}{x0^{2} \cdot x7^{2}})}{\frac{x5 \cdot x6^{2}}{x0^{2} \cdot x7}} + -5.79 \cdot \dfrac{tanh(\frac{x0^{2} \cdot x3 \cdot x4^{2}}{x5^{2} \cdot x7^{2}})}{cosh(\frac{x0^{2} \cdot x3 \cdot x4^{2}}{x5^{2} \cdot x7})^2}$ \\
\bottomrule
\end{tabular}
\end{table}

\section{Statistical Tests}

The following tables show the Bonferroni-adjusted $p$-values of a pairwise Wilcoxon signed rank test of the $RMSE$ scores for the test data of all the tested algorithms when compared to ITEA. The bold values highlight all the $p$-values $<0.05$.

\begin{landscape}
\begin{table}[ht]
\centering
\caption{Bonferroni-adjusted $p$-values of a pairwise Wilcoxon signed rank test of the $RMSE$ scores for the test data of each algorithm compared with ITEA.} 
\begin{tabular}{rllllllllll}
  \hline
 & Airfoil & Concrete & Cooling & Heating & Geo & Tecator & Tower & Red & White & Yacht \\ 
  \hline
  FEATFull & {\bf 5.9e-12} & 1.0e+00 & 6.8e-02 & 1.0e+00 & {\bf 3.8e-11} & 1.0e+00 & 1.0e+00 & 1.0e+00 & {\bf 1.5e-03} & {\bf 2.0e-02} \\ 
  FEAT & {\bf 1.5e-13} & {\bf 5.7e-04} & {\bf 4.2e-05} & {\bf 9.9e-08} & {\bf 1.5e-15} & 1.0e+00 & {\bf 6.7e-06} & 1.0e+00 & {\bf 4.0e-03} & {\bf 2.3e-09} \\ 
  SymTree & 5.4e-01 & 1.2e-01 & 1.0e+00 & {\bf 4.8e-04} & {\bf 4.8e-04} & 1.2e-01 & 1.0e+00 & 1.0e+00 & 1.0e+00 & 5.4e-02 \\ 
  GSGP & {\bf 1.3e-15} & {\bf 7.6e-12} & {\bf 9.0e-13} & {\bf 1.3e-15} & {\bf 1.3e-15} & 1.0e+00 & {\bf 1.3e-15} & 1.0e+00 & {\bf 1.7e-03} & {\bf 1.3e-15} \\ 
  GPLearn & {\bf 2.4e-09} & {\bf 2.4e-09} & {\bf 2.4e-09} & {\bf 2.4e-09} & {\bf 2.3e-09} & {\bf 4.0e-14} & {\bf 2.3e-09} & {\bf 1.7e-08} & {\bf 7.8e-07} & {\bf 3.2e-09} \\ 
  DCGP & {\bf 1.3e-15} & {\bf 1.3e-15} & {\bf 1.3e-15} & {\bf 1.3e-15} & {\bf 1.3e-15} & {\bf 1.3e-15} & {\bf 1.3e-15} & {\bf 2.4e-09} & {\bf 4.5e-09} & {\bf 2.4e-09} \\ 
  forest & 1.0e+00 & {\bf 3.6e-13} & {\bf 4.3e-08} & 1.3e-01 & {\bf 6.7e-13} & 1.0e+00 & {\bf 1.3e-15} & {\bf 4.4e-05} & {\bf 1.3e-15} & {\bf 1.1e-08} \\ 
  knn & {\bf 4.8e-04} & 5.4e-02 & {\bf 9.6e-04} & {\bf 4.8e-04} & 1.6e-01 & 1.0e+00 & {\bf 4.8e-04} & 1.0e+00 & 1.0e+00 & {\bf 4.8e-04} \\ 
  tree & {\bf 2.3e-07} & 1.0e+00 & {\bf 7.5e-09} & {\bf 1.6e-02} & 1.0e+00 & {\bf 2.2e-04} & {\bf 1.6e-10} & 5.8e-02 & {\bf 3.0e-04} & {\bf 6.0e-07} \\ 
  elnet & {\bf 4.8e-04} & {\bf 4.8e-04} & {\bf 4.8e-04} & {\bf 4.8e-04} & {\bf 4.8e-04} & 1.6e-01 & {\bf 4.8e-04} & 1.0e+00 & 6.5e-01 & {\bf 4.8e-04} \\ 
  lasso & {\bf 4.8e-04} & {\bf 4.8e-04} & {\bf 4.8e-04} & {\bf 4.8e-04} & {\bf 4.8e-04} & 3.0e-01 & {\bf 4.8e-04} & 1.0e+00 & 6.5e-01 & {\bf 4.8e-04} \\ 
  lassolars & {\bf 4.8e-04} & {\bf 4.8e-04} & {\bf 4.8e-04} & {\bf 4.8e-04} & {\bf 4.8e-04} & 1.6e-01 & {\bf 4.8e-04} & 1.0e+00 & 6.5e-01 & {\bf 4.8e-04} \\ 
  ridge & {\bf 4.8e-04} & {\bf 4.8e-04} & {\bf 4.8e-04} & {\bf 4.8e-04} & {\bf 4.8e-04} & 1.6e-01 & {\bf 4.8e-04} & 1.0e+00 & 6.5e-01 & {\bf 4.8e-04} \\ 
   \hline
\end{tabular}
\label{tab:pvals}
\end{table}
\end{landscape}

\end{document}